\pdfoutput=1
\documentclass[10pt]{article}
\usepackage[font={small,it}]{caption}
\usepackage{amsmath,amssymb,amsthm}
\usepackage{graphicx}
\usepackage{epsfig}
\usepackage[caption=false]{subfig}
\newcommand{\R}{\ensuremath{\mathbb{R}}}
\newcommand{\tr}{\mathop{\mathrm{Tr}}\nolimits}
\newcommand*{\defeq}{\mathrel{\vcenter{\baselineskip0.5ex \lineskiplimit0pt
                     \hbox{\scriptsize.}\hbox{\scriptsize.}}}
                     =}
\newcommand*{\eqdef}{=\mathrel{\vcenter{\baselineskip0.5ex \lineskiplimit0pt
                     \hbox{\scriptsize.}\hbox{\scriptsize.}}}
                     }
\newtheorem{thm}{Theorem}
\usepackage[pagebackref=false,breaklinks=false,letterpaper=true,bookmarks=false]{hyperref}

\date{}

\title{A Survey of Structure from Motion}
\author{Onur~\"{O}zye\c{s}il\footnotemark[2]
\and Vladislav Voroninski\footnotemark[3]
\and Ronen~Basri\footnotemark[4]
\and Amit~Singer\footnotemark[5]}

\begin{document}
\maketitle

\renewcommand{\thefootnote}{\fnsymbol{footnote}}

\footnotetext[2]{INTECH Investment Management LLC, One Palmer Square, Suite 441, Princeton, NJ 08542, USA ({\tt\small oozyesil@intechjanus.com}).}
\footnotetext[3]{Helm.ai, Menlo Park, CA 94025, USA ({\tt\small vlad@helm.ai}).}
\footnotetext[4]{Department of Computer Science and Applied Mathematics, Weizmann Institute of Science, Rehovot, 76100, ISRAEL ({\tt\small ronen.basri@weizmann.ac.il}).}
\footnotetext[5]{Department of Mathematics and PACM, Princeton University, Princeton, NJ 08544-1000, USA ({\tt\small amits@math.princeton.edu}).}

\renewcommand{\thefootnote}{\arabic{footnote}}

\begin{abstract}
The structure from motion (SfM) problem in computer vision is the problem of recovering the three-dimensional ($3$D) structure of a stationary scene from a set of projective measurements, represented as a collection of two-dimensional ($2$D) images, via estimation of motion of the cameras corresponding to these images. In essence, SfM involves the three main stages of (1) extraction of features in images (e.g., points of interest, lines, etc.) and matching these features between images, (2) camera motion estimation (e.g., using relative pairwise camera positions estimated from the extracted features), and (3) recovery of the $3$D structure using the estimated motion and features (e.g., by minimizing the so-called \textit{reprojection error}). This survey mainly focuses on relatively recent developments in the literature pertaining to stages (2) and (3). More specifically, after touching upon the early factorization-based techniques for motion and structure estimation, we provide a detailed account of some of the recent camera \textit{location} estimation methods in the literature, followed by discussion of notable techniques for $3$D structure recovery. We also cover the basics of the \textit{simultaneous localization and mapping (SLAM)} problem, which can be viewed as a specific case of the SfM problem. Further, our survey includes a review of the fundamentals of feature extraction and matching (i.e., stage (1) above), various recent methods for handling ambiguities in $3$D scenes, SfM techniques involving relatively uncommon camera models and image features, and popular sources of data and SfM software.  
\end{abstract}
\section{Introduction}
\label{sec:Intro}
\begin{figure}[!htbp]
\begin{center}
\includegraphics[trim=1cm 0cm 1cm 0cm, clip=true, width=0.8\linewidth]{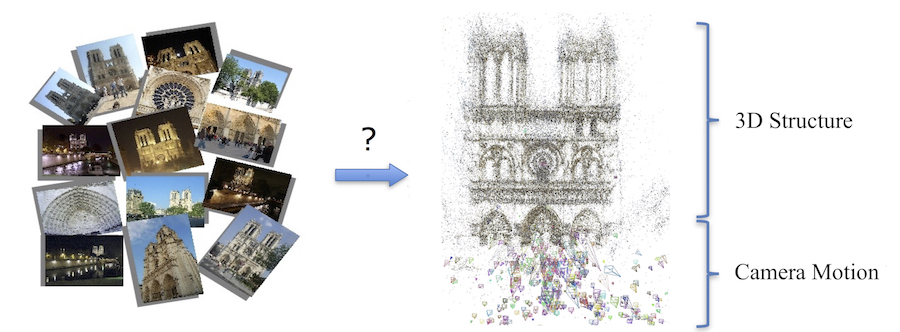}
\end{center}
\caption{The structure from motion (SfM) problem.\label{fig:SfMProblem}}
\end{figure}

Recovering the $3$-dimensional structure of a scene from images is a fundamental goal of computer vision. A particularly effective approach to $3$D reconstruction involves the use of many images of a stationary scene. This problem, commonly referred to as \textit{multiview structure from motion}  (SfM) (depicted in Figure~\ref{fig:SfMProblem}), is the subject of a large body of research in computer vision, starting with the seminal paper of Longuet-Higgins~\cite{LonguetHiggins}. A comprehensive and in-depth summary of this enormous body of work can be found in~\cite{HartleyBook}.

Modern methods usually solve the multiview SfM problem using \textit{bundle adjustment} techniques, which aim to optimize a cost function known as the \textit{total reprojection error} (cf.~\S\ref{sec:StructEst} for a more detailed and general discussion of bundle adjustment). With this cost function, given $n$ images of a stationary scene, the objective is to simultaneously determine the structure ($3$D coordinates of scene points) and the calibration parameters of each of the $n$ cameras that minimize the discrepancy between image measurements and their \textit{predictive model}. For instance, in the specific case of pairwise point correspondences between images, let us denote by $\{\mathbf{P}_j\}_{j=1}^p\subseteq\R^3$ the unknown positions of scene points, by $\{C_i\}_{i=1}^n$ the $3 \times 4$ camera matrices, where the entries of each $C_i$ encode the location of camera $i$, its orientation and internal calibration parameters. Also, let $(x_{ij},y_{ij})\in\R^2$ denote the measured projection of point $\mathbf{P}_j$ onto the image plane of camera $C_i$. Then, we can define a (relatively simple) bundle adjustment instance based on a sum of least squares cost function as
\begin{equation}
\label{eq:LSBA}
\underset{{\scriptstyle \{\mathbf{P}_j\}, \{C_i\}}}{\text{minimize}}  \ \ \sum_{i\sim j} \left(x_{ij} - \frac{C_{i1}^T\mathbf{P}_j}{C_{i3}^T\mathbf{P}_j} \right)^2 + \left(y_{ij} -  \frac{C_{i2}^T\mathbf{P}_j}{C_{i3}^T\mathbf{P}_j} \right)^2,
\end{equation}
where, $C_{ik}\in\R^4$ denotes the $k$'th row of $C_i$ ($1 \le k \le 3$), $i\sim j$ means that the $j$'th scene point is visible by the $i$'th camera, and we abuse notation and identify the representation $[\mathbf{P}_j, 1]\in\R^4$ of $\mathbf{P}_j$ in homogeneous coordinates by $\mathbf{P}_j$. Unfortunately, as a result of the special structure of the camera matrices and the cost function in (\ref{eq:LSBA}), the bundle adjustment problem (\ref{eq:LSBA}) is not convex, and its (na\"{i}ve) optimization in realistic settings typically converges to an (often undesired) local minimum. It is therefore critical to develop methods that can initialize bundle adjustment, i.e. that provide initial camera motion (and intrinsic calibration) and $3$D structure estimates, as closely as possible to the true solution.

In 2006 Snavely et al.~\cite{SnavelyData} presented a sequential pipeline for SfM, demonstrating that it can produce accurate reconstructions in practical scenarios where hundreds, or even thousands of independently captured photographs are provided, sparking a huge interest in the development of efficient SfM techniques for large, unordered image sets (cf.~\S\ref{sec:StructEst}). The suggested pipeline begins by detecting keypoints in each image. It then uses the SIFT descriptor~\cite{SIFT} to compare those keypoints across images and to produce a set of potential matches. Random sampling and consensus (RANSAC)~\cite{ransac} is applied next, to robustly estimate essential matrices between pairs of images (for computation of relative motion of camera pairs) and to discard outlier matches. Then, starting with a pair of images for which the largest number of inlier matches were found and then greedily adding one image at at a time, bundle adjustment is solved repeatedly. Although this sequential pipeline is computationally challenging, it successfully deals with large collections of images, producing in many cases highly accurate reconstructions. However, it is based on greedy steps that may not result in an optimal solution. Clearly, global approaches, which consider all images simultaneously (at least for the initial camera motion estimation), may potentially yield improved solutions. Indeed a number of recent methods attempt to globally estimate the camera locations (cf.~\S\ref{sec:CamLocEst}) and orientations. In this survey, for the initial camera motion estimation part of SfM, we focus on methods for camera location estimation in \S\ref{sec:CamLocEst}, and refer the reader to other surveys that review the problems of camera orientation and calibration recovery~\cite{TronSurvey}.

The outline of this survey is as follows. In \S\ref{sec:EarlyWorks}, we shortly discuss some of the early works in the literature that introduce the basic SfM framework and the relatively simple factorization methods. \S\ref{sec:CamLocEst} covers some of the recent work on camera location estimation from pairwise keypoint matches, such as the simpler least squares methods based on the (homogeneous) epipolar constraints, relatively stable convex methods based on a linearized representation (cf.~(\ref{eq:AltLinSys})) of pairwise direction measurements, heuristic methods for outlier rejection in pairwise directions, etc. We provide part of the key ideas and methods in the literature for $3$D structure estimation, with an emphasis on contemporary methods aimed at processing large, unordered sets of images, in \S\ref{sec:StructEst}. Details of influential works on the \emph{simultaneous localization and mapping} (SLAM) problem, which has recently experienced a dramatic increase of popularity, are discussed in \S\ref{sec:SLAM}. Some of the remaining crucial topics in the vast SfM literature, including methods for feature point extraction and matching, works on handling instabilities induced by symmetries and ambiguities in the images, methods for relatively uncommon types of cameras (e.g., omnidirectional cameras), techniques using alternative basic measurements (e.g., based on lines instead of feature points), some of the important software packages and popular datasets, etc., are considered in \S\ref{sec:Misc}. Lastly, we conclude with a discussion of the current state and possible future directions for the SfM problem in \S\ref{sec:Conclusion}.

As it is, rather unfortunately, the fate of any survey on a topic having an extensive body of literature like SfM, we are unable to cover every important work in the field. In essence, we try to focus on relatively recent SfM techniques. For earlier works in the literature, we refer the reader to other sources, e.g.~\cite{BundleAdjustment,HartleyBook,OliensisCritique}.

\section{Early Works}
\label{sec:EarlyWorks}
\begin{figure}[!htbp]
\begin{center}
\includegraphics[trim=0cm 0cm 0cm 0cm, clip=true, width=0.5\linewidth]{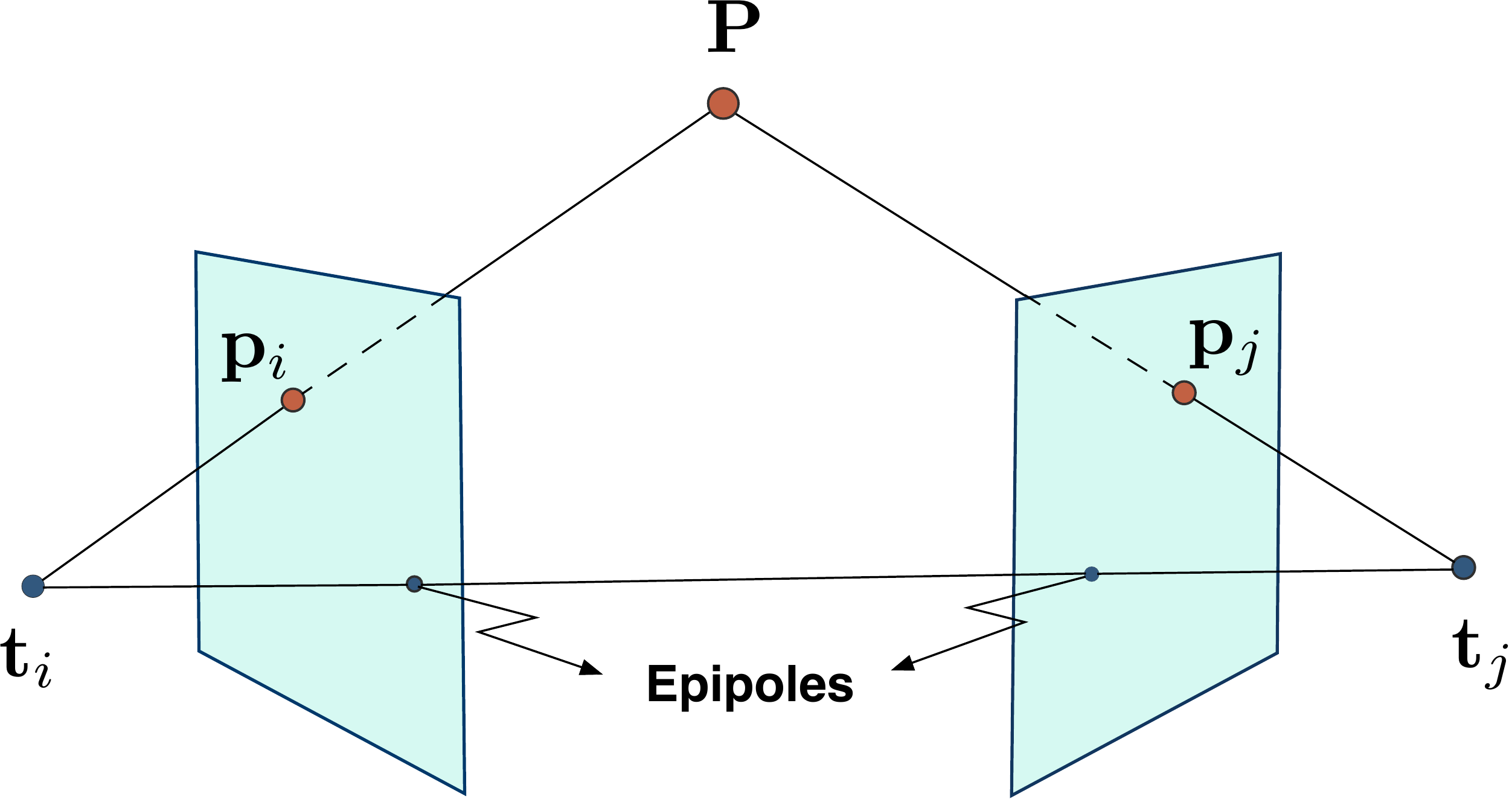}
\end{center}
\caption{The pinhole camera model \label{fig:EpipolarGeo}}
\end{figure}

\begin{figure}[!htbp]
\begin{center}
\includegraphics[trim=1.5cm 0cm 1.5cm 0cm, clip=true, width=0.8\linewidth]{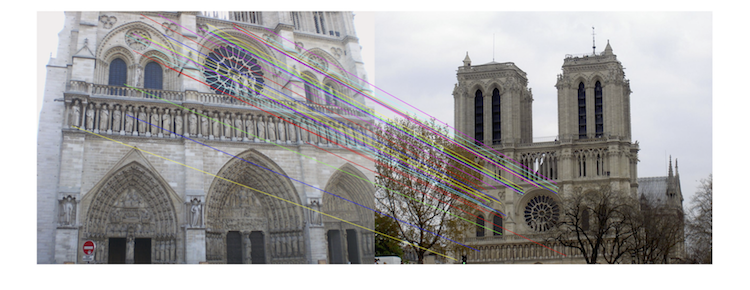}
\end{center}
\caption{Two images from the Notre-Dame Cathedral set in~{\rm\cite{SnavelyData}}, with corresponding feature points (extracted using the scale-invariant feature transform (SIFT)~{\rm\cite{SIFT}}).\label{fig:CorrPtsEx}}
\end{figure}

This section covers some of the relatively early works on SfM, which have introduced the basic problem formalism and the factorization based methods. We first consider the seminal work~\cite{LonguetHiggins} that introduced the first linear method based on point correspondences, later named \textit{the eight point algortihm}, to solve the SfM problem for a pair of cameras. Specifically, for a pair of cameras, \cite{LonguetHiggins} aims to estimate the relative camera motion, i.e.~relative rotation and translation, and the $3$D coordinates of the scene points captured by these cameras. Here, we represent the $i$'th camera using its orientation $R_i\in\mbox{SO}(3)$ and its location $\mathbf{t}_i\in\R^3$ by $(R_i, \mathbf{t}_i)$. Considering that we can always fix the (intrinsic) coordinate system of one of the two cameras to be the global coordinate system (i.e., we can set $R_i = I$ and $\mathbf{t}_i = \mathbf{0}$, since an absolute coordinate system cannot be determined from point correspondences), solving for the relative motion for a camera pair, and for the scene points based on the computed motion, then corresponds to actually solving the SfM problem for the pair. The problem setup\footnote{We note that, our notation and type of projective measurement (or, equivalently, the global coordinate system) are different from those of~\cite{LonguetHiggins}, even though both choices are equivalent (i.e., any result obtained by using one can be represented in terms of the other). Our choices (more or less) reflect the common terminology in the literature (cf., e.g., \cite{CvXSfM,LUD,MicaAmitSfM}). We also represent the epipolar constraints in their general form for a pair $(i,j)$ to make use of them in the following sections.} involves a simple pinhole camera model (also see, e.g., \S4 in~\cite{CvXSfM} or \cite{HartleyBook}), in which a scene point $\mathbf{P} \in \R^3$ is represented in the $i$'th image plane by $\mathbf{p}_i \in \R^3$ (as in Figure~\ref{fig:EpipolarGeo}).
Here, we obtain $\mathbf{p}_i$ by first representing $\mathbf {P}$, in the $i$'th camera's coordinate system, as $\mathbf {P}_i = R_i^T(\mathbf{P} - \mathbf{t}_i) = (\mathbf{P}_i^x,\mathbf{P}_i^y,\mathbf{P}_i^z)^T$ and then projecting it to the $i$'th image plane by $\mathbf{p}_i = (f_i/P_i^z)\mathbf{P}_i$, where $R_i\in\mbox{SO}(3), \mathbf{t}_i\in\R^3$ and $f_i\in\R^+$ denote the orientation, the location of the focal point and the focal length of the $i$'th camera, respectively. For a pair of cameras $i$ and $j$ as in Figure~\ref{fig:EpipolarGeo}, we can restate the coplanarity of the points $\mathbf{P},\mathbf{t}_i$ and $\mathbf{t}_j$, i.e. that $[(\mathbf{P} - \mathbf{t}_i)\times(\mathbf{P} - \mathbf{t}_j)]^T(\mathbf{t}_i-\mathbf{t}_j) = 0$, in terms of the observable corresponding point measurements $\mathbf{p}_i, \mathbf{p}_j$ (cf. Figure~\ref{fig:CorrPtsEx} for a depiction of a set of corresponding points between a pair of images) and the camera parameters as
\begin{equation}
\label{eq:EpipolarConst}
\mathbf{p}_i^T\left(\left[R_i^T(\mathbf{t}_{j}-\mathbf{t}_{i})\right]_{\times}R_{i}^TR_j\right)\mathbf{p}_j \eqdef \mathbf{p}_i^TE_{ij}\mathbf{p}_j  = 0 \ , \\
\end{equation}
where $[\mathbf{t}]_{\times}$ is the skew-symmetric matrix corresponding to the cross product with $\mathbf{t}$ and $E_{ij} = \left[R_i^T(\mathbf{t}_{j}-\mathbf{t}_{i})\right]_{\times}R_{i}^TR_j$ is \textit{the essential matrix} for the cameras $i$ and $j$. Also let $T_{ij}\defeq\left[R_i^T(\mathbf{t}_{j}-\mathbf{t}_{i})\right]_{\times}$ and $R_{ij} \defeq R_{i}^TR_j$, yielding $E_{ij} = T_{ij}R_{ij}$. \cite{LonguetHiggins} makes the crucial observation that, among the various equivalent restatements of the coplanarity of $\mathbf{P},\mathbf{t}_i$ and $\mathbf{t}_j$, the useful property of (\ref{eq:EpipolarConst}), which is known as \textit{the epipolar constraint}, is that it provides a basis for the estimation of the specially structured essential matrices in terms of the observable corresponding points. In other words, by fixing the undetermined scale for the entries of $E_{ij}$ (e.g., $\|E_{ij}\|_F = 1$, or as in~\cite{LonguetHiggins} $\|\mathbf{t}_{j}-\mathbf{t}_{i}\| = 1$, which is equivalent to $\|E_{ij}\|_F = \sqrt{2}$), we can solve for $E_{ij}\in\R^{3\times3}$ from eight (linearly independent) epipolar constraints (\ref{eq:EpipolarConst}) corresponding to eight $3$D points\footnote{In fact, due to the special structure of $E_{ij}$, it is well known (cf., e.g., \cite{HartleyBook}) that only five corresponding $3$D points are sufficient to estimate $E_{ij}$, however, this requires solving a nonlinear system of equations.}. Although \cite{LonguetHiggins} does not provide any algorithm for this purpose, nor does it consider the effects of uncertainties (i.e., noise) in the corresponding point measurements, the usual approach in the literature is simply to minimize the sum of squared errors in the epipolar constraints subject to the scale constraint, which is equivalent to finding the singular vector of the resulting data matrix corresponding to its smallest singular value. After estimating $E_{ij}$ (up to a sign to be determined later), \cite{LonguetHiggins} solves for $T_{ij}$ (up to another sign, again, to be determined later) by eliminating $R_{ij}$ using $E_{ij}E_{ij}^T = T_{ij}T_{ij}^T$. \cite{LonguetHiggins} then solves for $R_{ij}$, based on its orthogonality, and for the scene points using simple algebraic equations (cf.~\cite{LonguetHiggins} for details). Lastly, the undetermined signs of $E_{ij}$ and $T_{ij}$ are fixed by requiring the scene points to lie in front of both of the cameras. If the scene points are behind both of the cameras, the sign of $T_{ij}$ is altered and the relative rotation $R_{ij}$ and the scene points are recalculated. If the scene points fall behind one of the two cameras, the sign of $E_{ij}$ is altered and $T_{ij}, R_{ij}$ and the scene points are recalculated. Additionally,~\cite{LonguetHiggins} shortly discusses some of the degenerate cases of eight $3$D points (such as the cases of all points corresponding to the vertices of a cube, at least seven points lying on a plane, six of the eight points corresponding to the vertices of a regular hexagon, at least four points lying on a line), for which the eight point estimation cannot be used. We note that, although~\cite{LonguetHiggins} argues about the accuracy of the eight point algorithm, it was in fact observed to be sensitive to noise, which resulted in the construction of alternative methods, such as a normalized version of the eight point algorithm~\cite{HartleyEightPnt}, for relative pairwise motion estimation. For further reading on alternative methods to obtain epipolar geometries between images, cf., e.g., \cite{EpipolarReview,HartleyBook}.

Another early work in the literature that introduced the \textit{factorization method} for SfM is~\cite{TomasiKanadeFactor}. To model the SfM problem for objects that are relatively distant compared to their sizes, \cite{TomasiKanadeFactor} assumes an \textit{orthographic} camera model, in which the $3$D points are measured via parallel projections onto the image plane (consequently ignoring the camera translation along the optical axis). Let the location of the $j$'th $3$D point on the $j$'th camera plane be given by $\mathbf{q}_{ij} = (x_{ij},y_{ij})\in\R^2$. In the noiseless case, these orthographic measurements of $m$ $3$D points by $n$ cameras are represented in~\cite{TomasiKanadeFactor} in terms of a \textit{measurement matrix} $W\in\R^{2n\times m}$ satisfying
\begin{equation}
\label{eq:FactorMeasMatrix}
W =  \left[\begin{matrix} x_{11} & \hdots & x_{1m} \\ \vdots & \ddots & \vdots \\ x_{n1} & \hdots & x_{nm} \\ y_{11} & \hdots & y_{1m} \\ \vdots & \ddots & \vdots \\ y_{n1} & \hdots & y_{nm} \end{matrix}\right] \ .
\end{equation}
The crucial observation that~\cite{TomasiKanadeFactor} makes is that, if the origin of the global reference system is chosen to be at the center of the $3$D structure points (implying that each row of $W$ sums to zero), then $W$ can be decomposed into 
\begin{equation}
\label{eq:MeasMtxFactor}
W = RS \ , \ \mbox{for} \ R = \left[\begin{matrix}\mathbf{u}_1^T\\ \vdots \\ \mathbf{u}_n^T \\ \mathbf{v}_1^T\\ \vdots \\ \mathbf{v}_n^T \end{matrix}\right] \in \R^{2n\times3} \ \mbox{and} \ \ S = \left[\mathbf{P}_1 \hdots \mathbf{P}_m\right] \in\R^{3\times m} \ ,
\end{equation}
where $\mathbf{u}_k$ and $\mathbf{v}_k$ represent the orientation of the $k$'th camera (since, in the orthographic model, the third direction given by $\mathbf{u}_k\times\mathbf{v}_k$ is parallel to the viewing direction of the cameras) and $\mathbf{P}_k$ is the $k$'th $3$D point. As a result, $W$ has rank $3$. In the noisy case, which may result in a higher rank for $W$,~\cite{TomasiKanadeFactor} considers a valid measurement matrix to be given by the best rank-$3$ approximation to the noisy $W$ in the Frobenius norm sense, which is given by the singular value decomposition (SVD). Note that replacing $R$ and $S$ with $RQ$ and $Q^{-1}S$, for an arbitrary invertible matrix $Q\in\R^{3\times3}$ results in the same measurement matrix $W$. As a result, in order to compute $R$ and $S$ from $W$,~\cite{TomasiKanadeFactor} proposes to first compute a decomposition of $W$ using SVD, and then to solve for $Q$ using the orthonormality of the $n$ pairs $(\mathbf{u}_k,\mathbf{v}_k)$. That is, for $W = U_W\Sigma_WV_{W}^T$ representing the SVD of $W$, \cite{TomasiKanadeFactor} first lets $\hat{R} = U_W\Sigma^{1/2}$ and $\hat{S} = \Sigma^{1/2}V_{W}^T$, and then computes the solution $R = \hat{R}Q$ and $S = Q^{-1}\hat{S}$, where $Q$ is given to be a solution to the nonlinear system of $3n$ equations
\begin{equation}
\label{eq:FactorQ}
\hat{\mathbf{u}}_k^TQQ^T\hat{\mathbf{u}}_k = \hat{\mathbf{v}}_k^TQQ^T\hat{\mathbf{v}}_k = 1 \ , \ \hat{\mathbf{u}}_k^TQQ^T\hat{\mathbf{v}}_k = 0 \ , \ k = 1, \ldots, n
\end{equation}
where $\hat{\mathbf{u}}_k$ and $\hat{\mathbf{v}}_k$ represent the $k$'th and $(k+n)$'th rows of $\hat{R}$ (cf.~(\ref{eq:MeasMtxFactor})). \cite{TomasiKanadeFactor} also extends the factorization method to the case of \textit{occlusions} (i.e., to the case when all scene points are not visible by all of the cameras) and provides empirical evidence demonstrating the stability of their factorization approach on various experimental scenarios.

An extension of the factorization methodology for the multiview SfM with perspective cameras was introduced in~\cite{SturmFactorization}. This time, \cite{SturmFactorization} considers the basic image projection equations
\begin{equation}
\label{eq:FactorProjBasic}
\lambda_{ij}\mathbf{p}_{ij} = C_i\mathbf{P}_j \ , \ i = 1,\ldots,n \ , \ j = 1,\ldots,m \ ,
\end{equation}
where, for the $j$'th $3$D point represented (by abuse of notation) in the homogeneous coordinates by $\mathbf{P}_j\in\R^4$ and $C_i$ denoting the $i$'th camera matrix, $\mathbf{p}_{ij}\in\R^3$ is the projection of the $j$'th point onto the $i$'th camera plane (in the homogeneous coordinates) and $\lambda_{ij}$ are the undetermined scales, termed as \textit{projective depths}. Similar to~\cite{TomasiKanadeFactor}, these projective measurements are collected in~\cite{SturmFactorization} into a rank-$4$ measurement matrix $Y\in\R^{3n\times m}$ given by
\begin{equation}
\label{eq:FactorMeasMatrixProj}
Y =  \left[\begin{matrix} \lambda_{11}\mathbf{p}_{11} & \hdots & \lambda_{1m}\mathbf{p}_{1m}\\ \vdots & \ddots & \vdots \\ \lambda_{n1}\mathbf{p}_{n1} & \hdots & \lambda_{nm}\mathbf{p}_{nm} \end{matrix}\right] = \left[\begin{matrix}C_1\\ \vdots \\ C_n\end{matrix}\right]\left[\begin{matrix}\mathbf{P}_1 & \hdots  \mathbf{P}_m\end{matrix}\right] \ .
\end{equation}
The crucial point here is that we do not have direct access to the projective depths $\lambda_{ij}$, and if $\lambda_{ij}$ were to be accurately estimated, which constitutes the main contribution of~\cite{SturmFactorization}, a factorization technique similar to that of~\cite{TomasiKanadeFactor} could easily be applied. In order to achieve this goal, \cite{SturmFactorization} considers the linear system of equations for pairs of cameras $(i,j)$ given by  
\begin{equation}
\label{eq:FactorFndMtx}
F_{ij}\mathbf{p}_{jk}\lambda_{jk} = (\mathbf{e}_{ij}-\mathbf{p}_{ik})\lambda_{ik} \ ,
\end{equation}
where \textit{the fundamental matrix} $F_{ij}\in\R^{3\times3}$ satisfies $E_{ij} = K_{i}^TF_{ij}K_{j}$, for $K_{i}$ denoting the $i$'th camera calibration matrix and $E_{ij}$ the essential matrix for the pair $(i,j)$ (cf.~(\ref{eq:EpipolarConst})), and $\mathbf{e}_{ij}$ is the epipole on the $i$'th image plane (cf.~Figure~\ref{fig:EpipolarGeo}). Considering the homogeneous linear equations (\ref{eq:FactorFndMtx}) for the minimal number of $n-1$ camera pairs, \cite{SturmFactorization} estimates the projective depths up to global scale. After substitution of the recovered $\lambda_{ij}$ in (\ref{eq:FactorMeasMatrixProj}), \cite{SturmFactorization} proceeds similarly to~\cite{TomasiKanadeFactor} and factorizes $Y$ into its camera and structure components using SVD (however, this time, there is no need to compute an undetermined multiplier $Q$ since the camera and the structure parts of $Y$ do not have particular forms). Lastly, \cite{SturmFactorization} concludes with empirical results to evaluate the performance of their algorithm. A closely related approach, with an additional section for the extension of the algorithm in~\cite{SturmFactorization} to the case of lines, instead of points, as feature measurements is also given in~\cite{TriggsFactorization}.

An excellent account of factorization-based methods, for various camera models and alternative feature measurements, can also be found in the survey~\cite{KanadeFactorization}.

\section{Camera Location Estimation}
\label{sec:CamLocEst}
In this section we discuss parts of the existing literature focusing primarily on the camera location estimation part of the SfM problem. We consider the camera location estimation methods based on corresponding point estimates between pairs of images. In the majority of the methods we discuss, the corresponding point estimates are used to make relative motion measurements between pairs of images, which can be decomposed into pairwise rotational and translational measurements. The generic approach for these methods is to separately estimate the camera orientations based on the pairwise rotational measurements and to use these camera orientation estimates together with the pairwise translational measurements in order to solve for the camera locations. On the other hand, we also discuss some of the methods in the literature that, to some degree, deviate from the generic recipe mentioned above, and aim to jointly estimate camera orientations and locations, use local structure estimates for location estimation, consider triplets of cameras, etc.

In order to concretize the generic approach mentioned above, consider the simple pinhole camera model introduced in \S\ref{sec:EarlyWorks} and depicted in Figure~\ref{fig:EpipolarGeo}. As discussed in \S\ref{sec:EarlyWorks}, the essential matrix estimates, computed, e.g., using the epipolar constraints (\ref{eq:EpipolarConst}) in the presence of sufficiently many corresponding scene points (and the knowledge of the intrinsic camera parameters), can be uniquely factorized into relative rotational and translational parts, i.e. into estimates of $R_i^TR_j$ and $\left[R_i^T(\mathbf{t}_{j}-\mathbf{t}_{i})\right]_{\times}$. In general, the relative rotational parts are used for the estimation of camera orientations $R_i$ (cf.~\cite{TronSurvey} for a detailed account of the existing methods for camera orientation estimation). In order to estimate the camera locations, the estimated orientations can be used directly with the translational parts to obtain estimates of \textit{the pairwise directions} $\mathbf{\gamma}_{ij} = (\mathbf{t}_i-\mathbf{t}_j)/\|\mathbf{t}_i-\mathbf{t}_j\|$. An alternative approach is to go back to the epipolar constraints (\ref{eq:EpipolarConst}) to rewrite them as a set of linear homogeneous equations in the camera locations $\mathbf{t}_i$ given by
\begin{equation}
\label{eq:HomLinEqs}
(\mathbf{t}_i-\mathbf{t}_j)^T\mathbf{y}_{ij}^k = 0 \ , \ k = 1, \ldots, m_{ij}
\end{equation}
where $\mathbf{y}_{ij}^k$ are functions of $R_i, R_j$ and the $k$'th corresponding points $\mathbf{p}_i^k, \mathbf{p}_j^k$, and $m_{ij}$ denotes the number of corresponding points for the cameras $i$ and $j$. These equations can then be used directly for camera location estimation, or to obtain pairwise direction estimates.

A crucial point to note about the estimates of the pairwise directions $\gamma_{ij}$ is that they lack any \textit{scale information} pertaining to the distance $\|\mathbf{t}_i-\mathbf{t}_j\|$ between camera pairs. The difficulties\footnote{Note that the pairwise relative motion measurements are invariant also to global orientation and location shifts, i.e. any global rotational shift of the form $R_i\rightarrow RR_i$ and any global translation of the form $\mathbf{t}_i\rightarrow\mathbf{t}_i + \mathbf{t}$ for $R\in\mbox{SO}(3)$ and $\mathbf{t}\in\R^3$ produce equivalent solutions. However, contrary to the case of invariance to scale transformations of the form $\mathbf{t}_i\rightarrow c\mathbf{t}_i$ for $c\in\R^+$, these invariances are not observed to result in any difficulties in motion estimation.} arising from this homogeneity can be regarded as the most important contributor to the diversity of the camera location estimation methods in the literature.

A relatively direct approach for location estimation from pairwise directions is based on the homogeneous system of equations
\begin{equation}
\label{eq:DirHomEq}
\left(I-\mathbf{\gamma}_{ij}\mathbf{\gamma}_{ij}^T\right)(\mathbf{t}_i-\mathbf{t}_j) = \mathbf{0} \ ,
\end{equation}
which encapsulates the fact that the difference vectors $\mathbf{t}_i-\mathbf{t}_j$ are parallel to the pairwise directions $\mathbf{\gamma}_{ij}$. In~\cite{BATL2}, the locations are estimated by minimizing the \textit{sum of squared errors} in the system obtained by replacing $\mathbf{\gamma}_{ij}$ in (\ref{eq:DirHomEq}) with their estimates\footnote{In~\cite{BATL2}, the authors actually assume a slightly more general setup, where the directional measurements are not necessarily unit length, i.e. they may have some scale-related part (the source of which was unspecified). They also propose a \textit{maximum covariance spectral solution}, maximizing the alignment of $\mathbf{t}_i-\mathbf{t}_j$'s with $\mathbf{\gamma}_{ij}$'s, instead of penalizing their deviation. We do not discuss this method due to its inferior performance as reported by the authors of~\cite{BATL2}.} and using constraints to fix the global scale and translation, given respectively by $\sum_i \|\mathbf{t}_i\|^2 = 1$ and $\sum_i \mathbf{t}_i = \mathbf{0}$, to prevent the trivial solution of $\mathbf{t}_i \equiv \mathbf{t}$ for some $\mathbf{t}\in\R^3$, which is summarized as
\begin{equation}
\label{eq:LS}
\begin{aligned}
&\underset{{\scriptstyle \{\mathbf{t}_i\}}}{\text{minimize}}
& & \sum_{i\sim j}  (\mathbf{t}_i-\mathbf{t}_j)^T\left(I-\mathbf{\gamma}_{ij}\mathbf{\gamma}_{ij}^T\right)(\mathbf{t}_i-\mathbf{t}_j) \\
& \text{subject to}
& & \sum_{i} \|\mathbf{t}_i\|^2 = 1 \ ; \ \sum_{i} \mathbf{t}_i = \mathbf{0}
\end{aligned}
\end{equation}
Note that, considering the $n$ locations stacked into a vector $\mathbf{t}\in\R^{3n}$, whose $i$'th $3\times1$ block is equal to $\mathbf{t}_i$, we can rewrite the cost function of (\ref{eq:LS}) as $\mathbf{t}^TL\mathbf{t}$, for some appropriate $L\in\R^{3n\times3n}$, and the constraints as $\|\mathbf{t}\|^2=1$ and $V\mathbf{t} = \mathbf{0}$ (for $V = I_3\otimes\mathbf{1}_n^T$). Then, the solution to the least squares problem (\ref{eq:LS}) is given in terms of the (normalized) eigenvector corresponding to the fourth smallest eigenvalue of $L$, whose null space includes the three rows of $V$. The authors of~\cite{BATL2} also (heuristically) identify various classes of problem instances, which they refer to as ``problem pathologies'', resulting in relatively unstable solutions. Among the listed ``pathologies'', an important one is the class of ``underconstrained'' instances, which possess additional degrees of freedom resulting in more than one sets of location estimates (not related to each other via a global translation and/or scale transformation) that may be significantly different. As sources of these underconstrained instances the authors identify two (rather extreme) conditions, namely disconnectedness of the measurement graph\footnote{The measurement graph is composed of nodes for each camera $i$ and has an edge between nodes $i$ and $j$ if there is a direction estimate for this pair.} and the collinearity of all the directions at a single node, and also provide heuristic solutions to resolve the extra degrees of freedom in such instances (these two conditions are in fact only specific examples of a rich set of algebraic conditions defining underconstrained instances, as was fully characterized later in~\cite{CvXSfM,LUD}). A similar location estimation method based on the system of equations (\ref{eq:DirHomEq}) is given in~\cite{GovinduEarlyL2}. In this method, instead of directly minimizing the sum of squared errors in the system (\ref{eq:DirHomEq}), an iteratively reweighted least squares solver, where in the $k+1$'th iteration each equation in (\ref{eq:DirHomEq}) is weighted with $1/\|\mathbf{t}_i^k-\mathbf{t}_j^k\|$ for $\mathbf{t}_i^k$ denoting the solution of the $k$'th iteration, is used to democratize the contribution of each pair of cameras in the total cost function. Another approach, similar in spirit to these least squares methods, is introduced in~\cite{MicaAmitSfM}. This time, instead of the system of equations (\ref{eq:DirHomEq}), a least squares method is used to minimize the sum of squared errors in the epipolar constraints (\ref{eq:HomLinEqs}).

\begin{figure}[!htbp]
\begin{center}
\includegraphics[trim=11cm 9cm 16cm 15cm, clip=true, width=0.5\linewidth]{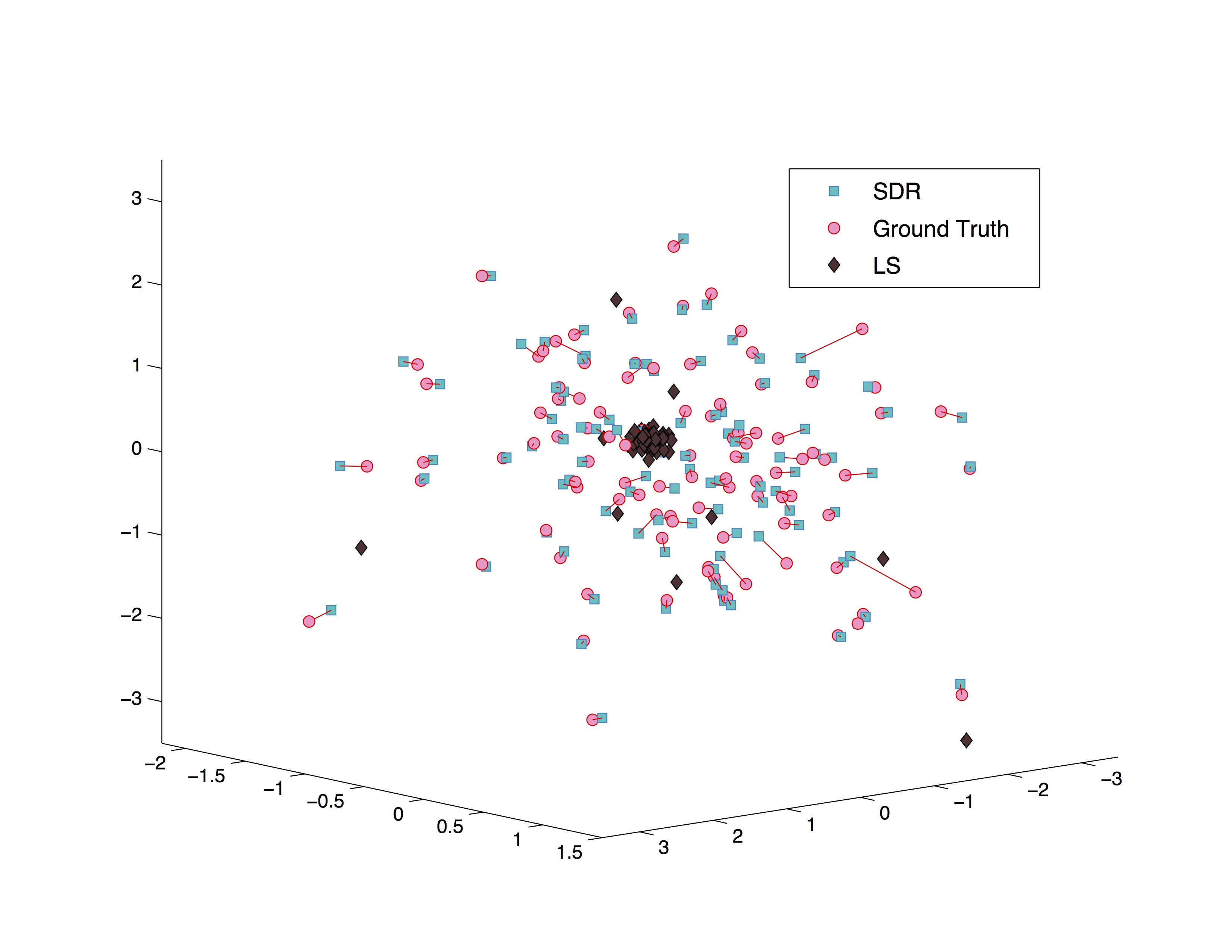}
\end{center}
\caption{Estimates for a noisy, synthetic instance of the camera location estimation problem for $100$ cameras, demonstrating the clustering estimates of the least squares (LS) method of~{\rm\cite{BATL2}} vs. estimates of the relatively stable semidefinite relaxation (SDR)  solver in~{\rm\cite{CvXSfM}} (the line segments represent the error incurred by the SDR solution compared to the ground truth).\label{fig:CollapseExample}}
\end{figure}
\begin{figure}[!htbp]
\begin{center}
\includegraphics[trim=0cm 0cm 0cm 0cm, clip=true, width=0.75\linewidth]{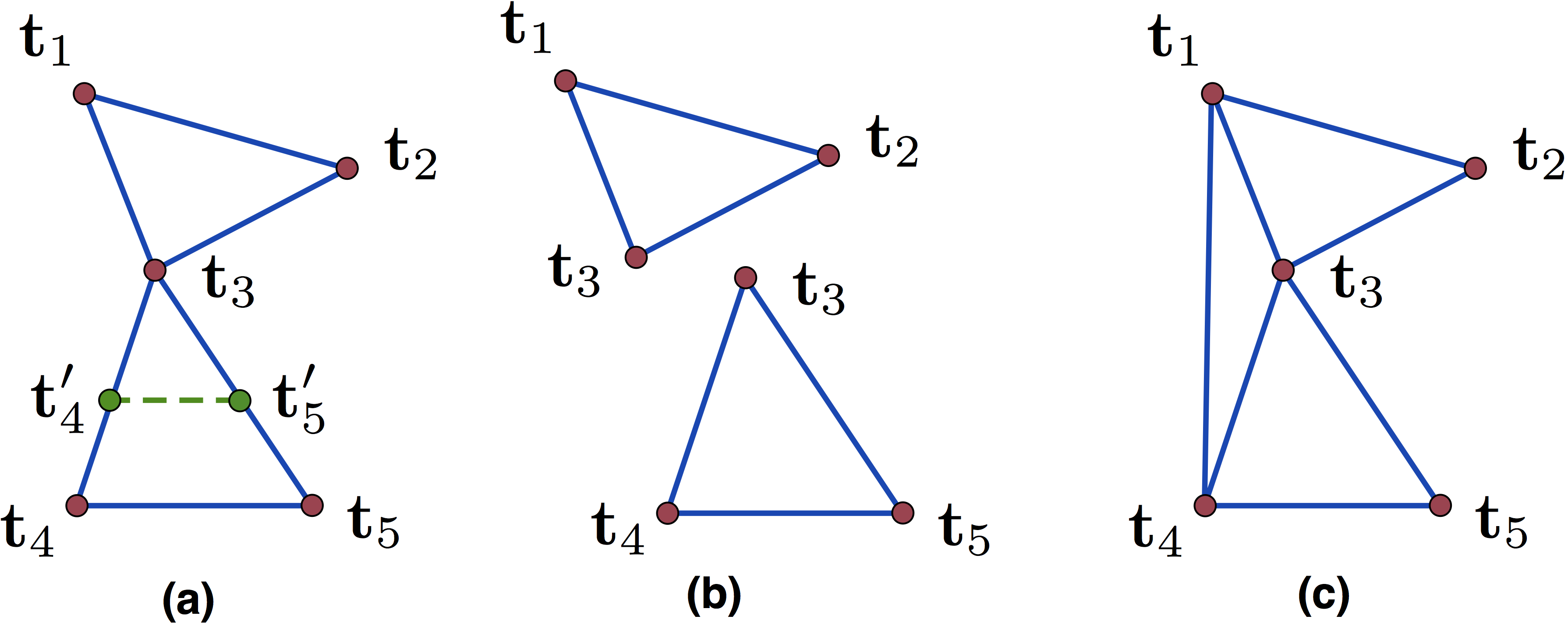}
\end{center}
\caption{{\rm(a)} A noiseless instance of the camera location estimation problem for $5$ locations on a connected graph (we assume noiseless direction measurements between location pairs having an edge), which is {\em not} parallel rigid (in $\R^2$ and $\R^3$). Non-uniqueness is demonstrated by two non-congruent location solutions $\{\mathbf{t}_1,\mathbf{t}_2,\mathbf{t}_3,\mathbf{t}_4,\mathbf{t}_5\}$ and $\{\mathbf{t}_1,\mathbf{t}_2,\mathbf{t}_3,\mathbf{t}_4',\mathbf{t}_5'\}$, each of which can be obtained from the other by an {\bf independent rescaling} of the solution for one of its maximally parallel rigid components, {\rm(b)} Maximally parallel rigid components of the formation in {\rm(a)}, {\rm(c)} A parallel rigid formation (in $\R^2$ and $\R^3$) obtained from the formation in {\rm(a)} by adding the extra edge $(1,4)$ linking its maximally parallel rigid components\label{fig:ParallelRigidityEx}}
\end{figure}

Even though the least squares methods of~\cite{BATL2,GovinduEarlyL2,MicaAmitSfM} are computationally very efficient and, for relatively small datasets, produce acceptable location estimates, the quality of their estimates degrades significantly for large, sparse, unordered and noisy datasets (i.e. for most of the real datasets studied in the literature). Typically, they tend to produce spurious location estimates that tightly cluster around a small number of points (cf. Figure~\ref{fig:CollapseExample} for such a clustering synthetic instance). This undesired tendency can be intuitively explained by considering the least squares optimization problem (\ref{eq:LS}), for which a clustering part of the solution results in a very small contribution to the cost function and a few locations, usually corresponding to nodes in the measurement graph having few edges, are separated from the cluster to satisfy the constraints. In order to resolve this difficulty, several different methods have been proposed in the literature. A relatively straightforward starting point, which was introduced in~\cite{CvXSfM}, is to require the locations to satisfy a maximal proximity condition by incorporating ``repulsion constraints'' of the form $\|\mathbf{t}_i-\mathbf{t}_j\|\geq c$, for some fixed $c>0$ (taken to be $c=1$ in~\cite{CvXSfM}), into the least squares formulation of~\cite{BATL2}. These non-convex constraints convert the least squares formulation into the notoriously difficult problem of 
\begin{equation}
\label{eq:NonConvexL2}
\begin{aligned}
&\underset{{\scriptstyle \{\mathbf{t}_i\}_{i\in V_t} \subseteq \R^d}}{\text{minimize}}
& & \sum_{(i,j)\in E_{t}}  \tr\left(\left(\mathbf{t}_i - \mathbf{t}_j\right)\left(\mathbf{t}_i - \mathbf{t}_j\right)^T\left(I-\mathbf{\gamma}_{ij}\mathbf{\gamma}_{ij}^T\right)\right)\\
&\text{subject to} 
& & \|\mathbf{t}_i - \mathbf{t}_j\|_2^2 \geq 1, \ \forall (i,j) \in E_t \ , \\
& & & \sum_i \mathbf{t}_i = \mathbf{0}
\end{aligned}
\end{equation}
To approximate this computationally difficult problem, \cite{CvXSfM} first introduces the rank-$1$, positive semidefinite matrix variable $T\in\R^{3n\times3n}$ (for $n$ denoting the number of cameras), the $ij$'th $3\times3$ block $T_{ij}$ of which is given by $T_{ij} = \mathbf{t}_i\mathbf{t}_j^T$. Using $T$, the cost function and the non-convex constraints of (\ref{eq:NonConvexL2}) are given, respectively, as linear functions $\sum_{i\sim j} \tr\left(\mathcal{L}_{ij}(T)\left(I-\mathbf{\gamma}_{ij}\mathbf{\gamma}_{ij}^T\right)\right)$ and $\tr\left(\mathcal{L}_{ij}(T)\right)\geq 1$ of $T$, where the operator $\mathcal{L}_{ij}(.)$ satisfies
\begin{equation}
\label{eq:LijT} 
\mathcal{L}_{ij}(T) = T_{ii} + T_{jj} - T_{ij} - T_{ji} = \left(\mathbf{t}_i - \mathbf{t}_j\right)\left(\mathbf{t}_i - \mathbf{t}_j\right)^T \ .
\end{equation}
As a result of the linearity in $T$ of the cost function and the constraints, the non-convexity of (\ref{eq:NonConvexL2}) is fully represented by the non-convex rank-$1$ constraint on $T$ (note that positive semidefiniteness is a convex constraint). As a result, \cite{CvXSfM} removes this final non-convex constraint to obtain the \textit{semidefinite relaxation} formulation given by
\begin{equation}
\label{eq:SDR}
\begin{aligned}
& \underset{{\scriptstyle T}}{\text{minimize}}
& & \sum_{i\sim j} \tr\left(\mathcal{L}_{ij}(T)\left(I-\mathbf{\gamma}_{ij}\mathbf{\gamma}_{ij}^T\right)\right)\\
& \text{subject to} 
& & \tr\left(\mathcal{L}_{ij}(T)\right)\geq 1\ , \ \forall i\sim j \ ,\\
& & & \tr\left(VT\right) = 0 \ , \\
& & & T \succeq 0 \ .
\end{aligned}
\end{equation}
The approximate solution $\mathbf{t}_i$ in (\ref{eq:NonConvexL2}) is obtained to be the $i$'th $3\times1$ block of the leading eigenvector of the solution of the SDR (\ref{eq:SDR}). The tightness\footnote{Tightness here refers to obtaining the solution of the non-convex problem as the solution to the relaxed convex problem, i.e. the two globally optimal solutions are the same and hence there is no \textit{relaxation gap}.} of this relaxation up to relatively high levels of noise is empirically demonstrated in~\cite{CvXSfM}, and also a \textit{stability of recovery} result is proven, which quantifies the amount of distortion in the location estimates to be on the order of the noise present in the pairwise directions, under fairly general assumptions for the noise model. In order to improve the quality of their location estimates, \cite{CvXSfM} proposes a ``robust pairwise direction estimation'' method based on the system (\ref{eq:HomLinEqs}), where $\mathbf{y}_{ij}^k$'s are viewed as noisy samples from the $2$D subspace orthogonal to $\mathbf{\gamma}_{ij}$ and efficient robust subspace recovery techniques are used to estimate each pairwise direction $\mathbf{\gamma}_{ij}$. Additionally, and perhaps more importantly, a complete characterization of the \textit{well-posed instances} of the location recovery from pairwise directions\footnote{To be more accurate, in~\cite{CvXSfM}, the information assumed to be available is actually a collection of potentially incomplete and noisy versions of the \textit{pairwise lines} (or unsigned directions), which can be represented by the projection matrices $(\mathbf{t}_i-\mathbf{t}_j)(\mathbf{t}_i-\mathbf{t}_j)/\|\mathbf{t}_i-\mathbf{t}_j\|^2$. However, the well-posed instances for this case turns out to be the same as in the case of pairwise directions.} is provided in~\cite{CvXSfM} by demonstrating its equivalence to the existing results in the field of \textit{parallel rigidity theory}. For instance, while camera orientation estimation only requires the connectivity of the measurement graph, connectivity alone is insufficient to make a camera location estimation instance well-posed (cf.~Figure~\ref{fig:ParallelRigidityEx} for such an instance).
The complete characterization of well-posed instances of the location recovery problem (for arbitrary dimensions $d\geq2$) , via (generically) parallel rigid measurement graphs is summarized in Theorem~\ref{thm:LamanConds}.
\begin{thm}
\label{thm:LamanConds}
A graph $G = (V,E)$ is generically parallel rigid in $\R^d$ if and only if it contains a nonempty set of edges $E'\subseteq E$, with $(d-1)|E'| = d|V| - (d+1)$, such that for all subsets $E''$ of $E'$, we have
\begin{equation}
\label{eq:LamanIneqs}
(d-1)|E''| \leq d|V(E'')| - (d+1) \ ,
\end{equation}
where $V(E'')$ denotes the vertex set of the edges in $E''$.
\end{thm}
The authors of~\cite{CvXSfM} also provide efficient algorithms for testing well-posedness, and algorithms for extracting maximal well-posed subproblems in the presence of an ill-posed instance. Additionally, \cite{CvXSfM} introduces an efficient \textit{alternating direction augmented Lagrangian method (ADM)} to solve the SDR (\ref{eq:SDR}), and a distributed algorithm to apply the SDR method for large sets of images. 

\begin{figure}[!htbp]
\begin{center}
\includegraphics[trim=0cm 0cm 0cm 0cm, clip=true, width=0.95\linewidth]{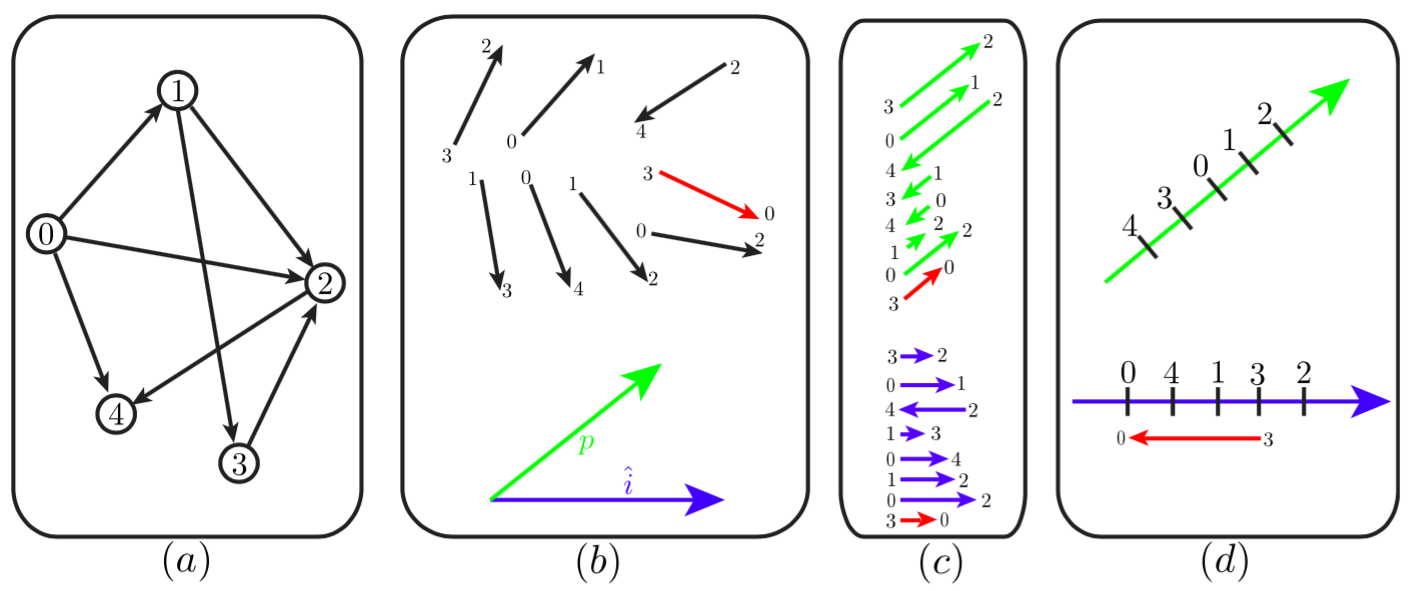}
\end{center}
\caption{A simplified illustration (courtesy of~{\rm\cite{Snavely1D}}) of outlier detection via 1DSfM in~{\rm\cite{Snavely1D}}: {\rm(a)} Ground truth camera locations with (potentially noisy) pairwise directions available for camera pairs connected with arrows {\rm(b)} Available pairwise directions, with an outlier edge shown in red, and two directions, $\hat{i}$ and $p$, for 1D projections {\rm(c)} Projections of pairwise directions onto $\hat{i}$ and $p$, resulting in two instances of the minimum feedback arc set problem {\rm(d)} Solutions to the 1D problems in {\rm(c)}, depicting the violation (satisfaction) of the 1D constraints by the outlier edge $(3,0)$ in the 1D problem induced by $\hat{i}$ ($p$), since outlier edges may be consistent with only some of the solutions to the 1D subproblems.\label{fig:Snavely1DSfM}}
\end{figure}

An alternative two step procedure, which is composed of the detection of outliers among the pairwise direction measurements, via a procedure named \textit{1DSfM}, followed by a (non-convex) location estimation method, is studied in~\cite{Snavely1D} (cf.~Figure~\ref{fig:Snavely1DSfM}, courtesy of~\cite{Snavely1D}, for a simplified illustration of the outlier detection procedure 1DSfM). 
The main idea in the outlier detection step, is to project all of the pairwise directions onto a one-dimensional (random) subspace and to try to solve the resulting ordering problem\footnote{This ordering problem, namely \textit{the minimum feedback arc set problem}, is known to be an NP-complete problem, hence \cite{Snavely1D} uses a heuristic/approximate method observed to perform well in the literature.} as consistently as possible, from which potential outliers can be identified as having larger inconsistencies. The detection procedure ``solves'' several of these randomized one-dimensional problems and identifies the outliers as those directions having an inconsistency score larger than some value. After removal of outlier direction measurements, the locations are estimated by an unconstrained maximization, using the Levenberg-Marquardt algorithm, of the non-convex cost function $\sum_{ij} \hat{\mathbf{\gamma}}_{ij}^T(\mathbf{t}_i-\mathbf{t}_j)/\|\mathbf{t}_i-\mathbf{t}_j\|$ that measures the alignment of the pairwise directions $(\mathbf{t}_i-\mathbf{t}_j)/\|\mathbf{t}_i-\mathbf{t}_j\|$ with their ``cleaned'' estimates $\hat{\mathbf{\gamma}}_{ij}$. Extensive experimental results on large datasets demonstrating the accuracy and the efficiency of their method are also provided in~\cite{Snavely1D}.

Instead of the systems of equations (\ref{eq:HomLinEqs}) or (\ref{eq:DirHomEq}), on which the various least squares methods are based, an alternative approach aiming to prevent the clustering phenomena is to consider the system of equations
\begin{equation}
\label{eq:AltLinSys}
\mathbf{t}_i-\mathbf{t}_j -\alpha_{ij}\mathbf{\gamma}_{ij} = \mathbf{0} \ ,
\end{equation}
where $\alpha_{ij} \defeq \|\mathbf{t}_i-\mathbf{t}_j\|$. Note that, if we ignore this defining (non-convex) relation between $\alpha_{ij}$'s and $\mathbf{t}_i$'s, (\ref{eq:AltLinSys}) turns out to be a linear system in $\alpha_{ij}$'s and $\mathbf{t}_i$'s for given $\mathbf{\gamma}_{ij}$. This system forms the basis of the location estimation method used in~\cite{TronVidalJournal}, where the cost function is the sum of squares of the errors in (\ref{eq:AltLinSys}) and the additional constraints $\alpha_{ij}\geq1$ are used to prevent the trivially collapsing solution $\mathbf{t}_i \equiv \mathbf{0}, \alpha_{ij}\equiv 0$. That is, for location estimation~\cite{TronVidalJournal} solves
\begin{equation}
\label{eq:CLS}
\begin{aligned}
&\underset{{\scriptstyle \{\mathbf{t}_i\}, \{\alpha_{ij}\}}}{\text{minimize}}
& & \sum_{i\sim j}  \left\| \mathbf{t}_i-\mathbf{t}_j - \alpha_{ij}\mathbf{\gamma}_{ij} \right\|^2 \\
& \text{subject to}
& & \sum_{i} \mathbf{t}_i = \mathbf{0} \ ; \ \alpha_{ij}\geq 1
\end{aligned}
\end{equation}
In fact, the main theme in~\cite{TronVidalJournal} is to provide a distributed algorithm for camera motion estimation, i.e. joint estimation of orientations and locations. In this respect, \cite{TronVidalJournal} provides a \textit{consensus} based three step algorithm, and local convergence and partial stability (for the rotational part) results. The quadratic location estimation method (\ref{eq:CLS}) is used to obtain an initial set of locations (i.e. constitutes the second step of their distributed algorithm), and then is fused with the rotational part to obtain a joint motion estimation method. In terms of the camera location estimation problem with given pairwise directions, the usage of the system (\ref{eq:AltLinSys}) as an alternative to the systems (\ref{eq:HomLinEqs}) or (\ref{eq:DirHomEq}) together with the relaxed repulsion constraints $\alpha_{ij}\geq1$ is the crucial contribution of~\cite{TronVidalJournal}, since, as reported, e.g., by~\cite{LUD}, this framework provides location estimates with acceptable quality, which do not tend to cluster for large and noisy datasets, while preserving computational efficiency. A closely related optimization method based on (\ref{eq:AltLinSys}) and the constraints $\alpha_{ij}\geq1$ is introduced in~\cite{MoulonLinfty}. The cost function of this method is given by the \textit{maximum absolute deviation} in the system (\ref{eq:AltLinSys})\footnote{The method of~\cite{MoulonLinfty} actually assumes the possibility of multiple pairwise direction measurements for each camera pair $i$ and $j$, computed from each triangle the pair is in.}. The authors of~\cite{MoulonLinfty} also introduce robust rotation and pairwise direction estimation methods. We note, however, that such maximal deviation penalty methods (i.e., $\ell_{\infty}$ norm minimization methods) are highly sensitive to outliers in the pairwise direction estimates.

\subsection{Robust Convex Methods}
The existence of large proportion of outliers among pairwise direction estimates, e.g.~stemming from systematically mismatched features between images as a result of symmetries or ambiguities in the images (also see \S\ref{subsec:Ambiguities}), manifests itself as large errors in the estimated locations. While for some datasets this difficulty may be tamed by preprocessing of the pairwise directions for rejection of outliers (e.g., as in~\cite{Snavely1D}), making use of consistencies in locally constructed $3$D structures, etc., generally applicable efficient algorithms, which are resilient to large number of outliers and which exhibit provable convergence to globally optimal solutions and admit theoretical guarantees of robustness, are of high value. We will consider two recent algorithms in this section having some of these desired properties.

\begin{figure}[!htbp]
\begin{center}
\includegraphics[trim=2.3cm 8cm 0.8cm 1.5cm, clip=true, width=0.65\linewidth]{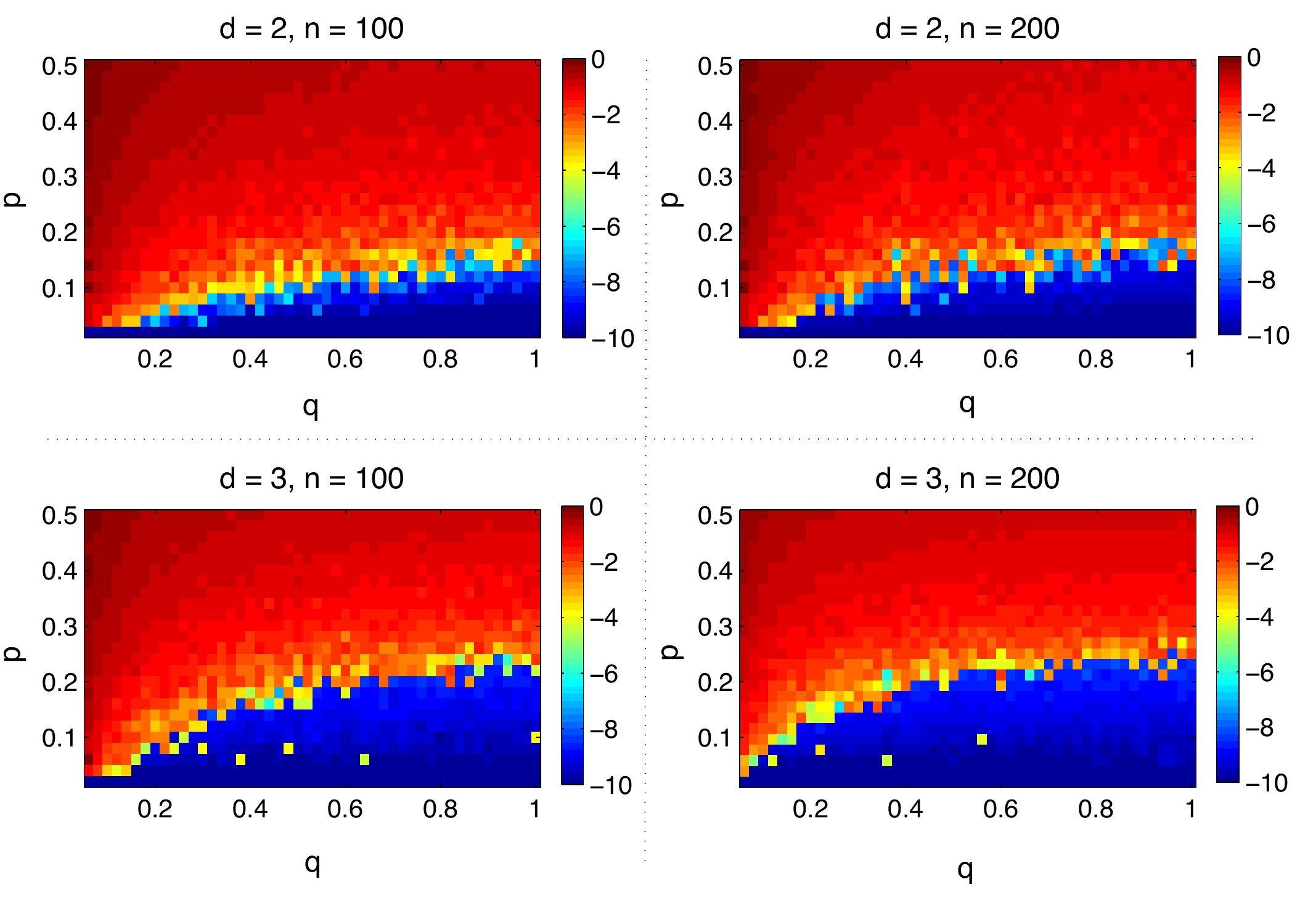}
\end{center}\vspace{-0.15in}
 \caption{Recovery error of the LUD solver~{\rm(\ref{eq:LUD})}, demonstrating exact recovery regimes in the presence of outliers for dimensions $d=2, 3$ and number of cameras $n=100, 200$. Measurement graphs, with an edge between cameras having pairwise directions, are sampled from the Erd\"{o}s-R\'{e}nyi ensemble. Camera locations are i.i.d.~Gaussian samples, independent of the measurement graph. The color intensity of each pixel represents the logarithmic recovery error $\log_{10}(\mbox{{\rm NRMSE}})$ (cf.~equation {\rm(14)} in~{\rm\cite{LUD}} for the definition of the normalized root mean squared error {\rm(NRMSE)}), depending on the probability $q$ of the existence of an edge for each pair ($x$-axis), and the probability $p$ of each existing edge to be an outlier ($y$-axis). {\rm NRMSE} values are averaged over $10$ trials.\label{fig:ExactRecoveryTest}}
\end{figure}

The first location estimation algorithm, based on the system (\ref{eq:AltLinSys}) and closely related to~\cite{TronVidalJournal,MoulonLinfty}, is given in~\cite{LUD}, where this time, instead of using a cost function given as the sum of squared errors (as in~\cite{TronVidalJournal}) or the maximum error (as in~\cite{MoulonLinfty}) in the system of equations (\ref{eq:AltLinSys}), the method minimizes the sum of \textit{unsquared} errors in (\ref{eq:AltLinSys}) (hence the name \textit{least unsquared deviations (LUD) method}) by employing the same relaxed repulsion constraints $\alpha_{ij}\geq1$. In other words, $\mathbf{t}_i$ are estimated by solving
\begin{equation}
\label{eq:LUD}
\begin{aligned}
&\underset{{\scriptstyle \{\mathbf{t}_i\}, \{\alpha_{ij}\}}}{\text{minimize}}
& & \sum_{i\sim j}  \left\| \mathbf{t}_i-\mathbf{t}_j - \alpha_{ij}\mathbf{\gamma}_{ij} \right\| \\
& \text{subject to}
& & \sum_{i} \mathbf{t}_i = \mathbf{0} \ ; \ \alpha_{ij}\geq 1
\end{aligned}
\end{equation}
The motivation in this formulation (mainly inspired by compressed sensing methods designed to recover corrupted signals from a minimal set of linear measurements), is to maintain robustness to outliers in the measurements of the pairwise directions $\mathbf{\gamma}_{ij}$. As a convex \textit{second-order cone program (SOCP)}, the LUD solver (\ref{eq:LUD}) is a computationally efficient location estimator with guaranteed convergence to the globally optimal solution. More importantly, the rather interesting phenomenon of exact location recovery\footnote{Here, ``exact location recovery'' corresponds to the estimation of locations up to a global scale and translation (i.e., up to a gauge freedom of kind $\mathbf{t}_i\rightarrow c\mathbf{t}_i + \mathbf{t}$ for arbitrary $c>0$ and $\mathbf{t}\in\R^3$), whose ground truth values are known, with a total error smaller than a fixed value, which may be chosen to represent machine precision errors.} by the LUD method in the presence of sufficiently many exact direction measurements (i.e. sufficiently few outlier direction measurements), the number of which varies with the level of sparsity in the measurement graph and the number of cameras, is empirically demonstrated for the first time  in~\cite{LUD} (cf.~Figure~\ref{fig:ExactRecoveryTest}). 
An efficient iteratively weighted least squares (IRLS) solver for the LUD problem (\ref{eq:LUD}) is provided in~\cite{LUD}. 

\begin{figure}[!htbp]
\centering
\includegraphics[trim=0.7cm 1cm 0.6cm 1.5cm, clip=true, width=0.65\linewidth]{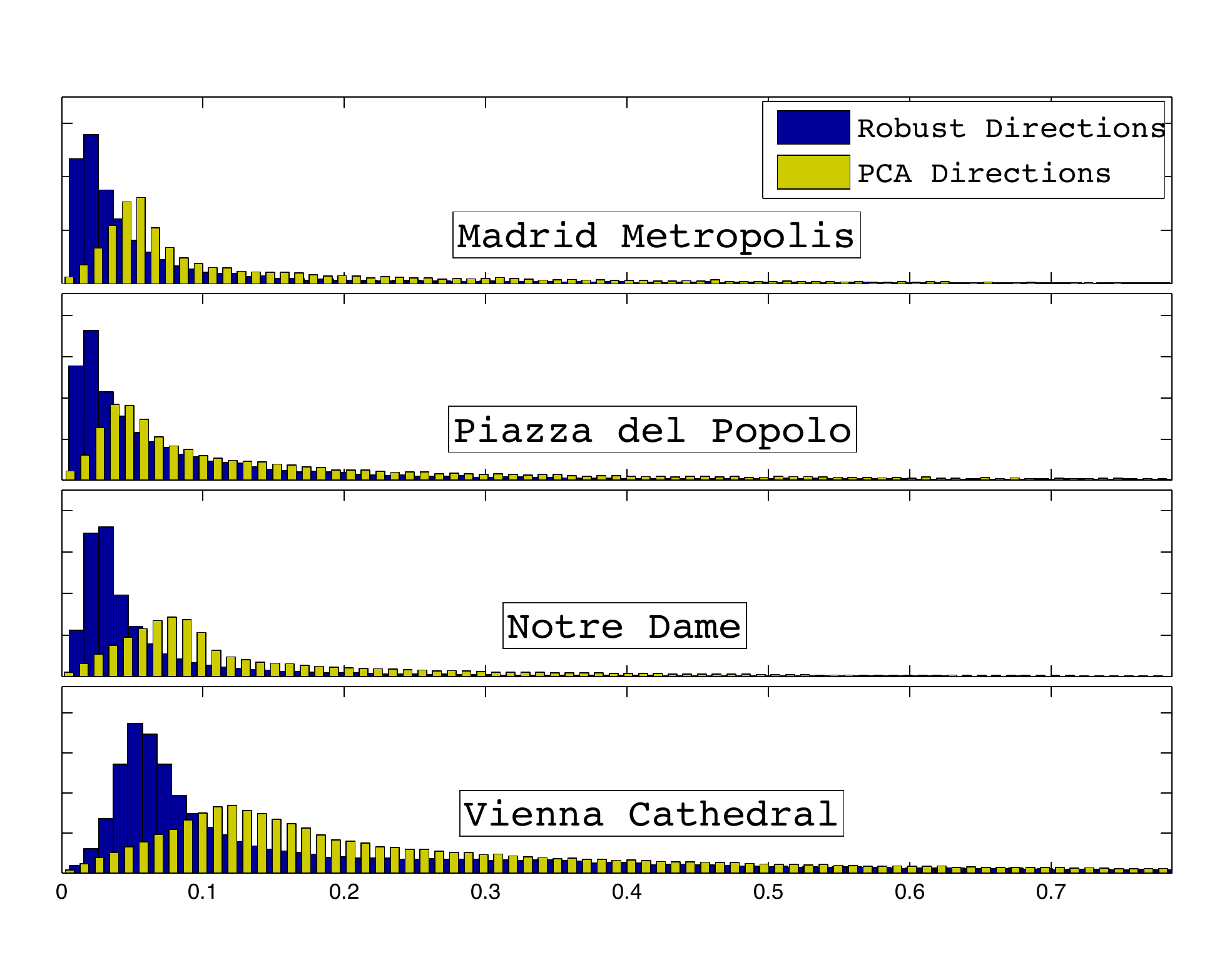}
\caption{Histogram plots of the errors in direction estimates computed by the robust method {\rm(\ref{eq:REAPER})} and the PCA method, which replaces the cost function in {\rm(\ref{eq:REAPER})} with the sum of squares version $\sum_{k=1}^{m_{ij}}  ((\gamma_{ij}^0)^T\hat{\mathbf{y}}_{ij}^k)^2$, for some of the datasets studied in~{\rm\cite{LUD}}. The errors represent the angles between the estimated directions and the corresponding ground truth directions (computed from a sequential SfM method based on~{\rm\cite{SnavelyData}}, and provided in~{\rm\cite{Snavely1D}}) (the errors take values in $[0,\pi]$, yet the histograms are restricted to $[0,\pi/4]$ to emphasize the difference of the quality in the estimated directions).\label{fig:DirEstimCompare}}
\end{figure}

Additionally, \cite{LUD} introduces a (non-convex) direction estimation algorithm based on the system (\ref{eq:HomLinEqs}) (similar to~\cite{CvXSfM}, but more efficient). The pairwise directions are estimated in a two step procedure that first produces estimates of unsigned directions $\mathbf{\gamma}_{ij}^0 = b_{ij}\mathbf{\gamma}_{ij}$, and then computes the initially undetermined signs $b_{ij}\in\{-1,+1\}$ via the chirality requirement on the $3$D points. In order to maintain robustness to outliers among estimates $\hat{\mathbf{y}}_{ij}^k$ of $\mathbf{y}_{ij}^k$ in (\ref{eq:HomLinEqs}) for the estimation of unsigned directions $\mathbf{\gamma}_{ij}^0$, \cite{LUD} considers the following (non-convex) problem:
\begin{equation}
\label{eq:REAPER}
\begin{aligned}
\underset{{\scriptstyle \gamma_{ij}^0}}{\text{minimize}}
& \ \ \sum_{k=1}^{m_{ij}}  | (\gamma_{ij}^0)^T\hat{\mathbf{y}}_{ij}^k| \\
\text{subject to} & \ \ \|\gamma_{ij}^0\| = 1 \,
\end{aligned}
\end{equation}
To obtain the estimate $\hat{\mathbf{\gamma}}_{ij}^0$, a (heuristic) IRLS method is used, which is not guaranteed to converge to global optima (since the program (\ref{eq:REAPER}) is not convex). However, \cite{LUD} provides empirical evidence for the robustness of the solution to (\ref{eq:REAPER}), while preserving computational efficiency (cf.~Figure~\ref{fig:DirEstimCompare}). 
Finally, \cite{LUD} demonstrates the high quality and the efficiency of the overall method on large datasets from~\cite{Snavely1D}. 

The excellent empirical performance of the LUD method has recently inspired the development of a new method for location estimation from pairwise directions called \textit{ShapeFit}~\cite{ShapeFit}. ShapeFit is based on a convex program that minimizes the sum of the \textit{unsquared} errors in the system (\ref{eq:DirHomEq}) subject to the constraint $\sum_{ij}(\mathbf{t}_i-\mathbf{t}_j)^T\mathbf{\gamma}_{ij} = 1$, which fixes the scale and sign of the location estimates, as well as encouraging correlation with the true relative directions. Specifically, the ShapeFit program consists of the following
\begin{equation}
\label{eq:ShapeFit}
\begin{aligned}
&\underset{{\scriptstyle \{\mathbf{t}_i\} }}{\text{minimize}}
& & \sum_{i\sim j}  \| (I-\mathbf{\gamma}_{ij}\mathbf{\gamma}_{ij}^T)(\mathbf{t}_i - \mathbf{t}_j) \|_2 \\
& \text{subject to}
& & \sum_{i} \mathbf{t}_i = \mathbf{0} \ ; \ \sum_{i\sim j}  (\mathbf{t}_i - \mathbf{t}_j)^T\gamma_{ij} = 1
\end{aligned}
\end{equation}
Similar to the LUD method of~\cite{LUD}, the motivation behind the cost function of ShapeFit is to improve robustness to outliers among the pairwise direction measurements. Also, ShapeFit is amenable to theoretical analysis, and the results of~\cite{ShapeFit} provide the first rigorous results guaranteeing exact location recovery from corrupted relative direction observations. More concretely, Theorem 1 in~\cite{ShapeFit} considers a high dimensional version of the location recovery problem (i.e., $\mathbf{t}_i\in\R^d$ for sufficiently large $d$) and proves exact location recovery with high probability for i.i.d.~Gaussian locations in the presence of a measurement graph sampled from an Erd\"{o}s-R\'{e}nyi ensemble for which at most a constant fraction of direction measurements for any particular location are arbitrarily corrupted. For the same setup, with the exception of requiring the fraction of corrupted directions measurements for any particular location to be, \emph{up to poly-logarithmic factors}, at most a constant, Theorem 2 of~\cite{ShapeFit} proves exact location recovery for the physically relevant case $d=3$. The physically relevant theorem is as follows
\begin{thm} 
\label{thm:ShapeFit3D} 
There exists $n_0 \in \mathbb{N}$ and $c \in \R$ such that the following holds for all $n \ge n_0$. Let the measurement graph $G([n],E)$ be drawn from the Erd\"{o}s-R\'{e}nyi ensemble $G(n,p)$ for some $p = \Omega( n^{-1/5} \log^{3/5} n)$. Let $\mathbf{t}^{(0)}_1, \ldots \mathbf{t}^{(0)}_n  \in \R^3$ denote the ground truth location, where $\mathbf{t}^{(0)}_i \sim \mathcal{N}(0, I_{3 \times 3})$ are i.i.d., independent from $G$. There exists $\gamma = \Omega(p^5 / \log^3 n)$ and an event of probability at least  $1- \frac{1}{n^{4}}$ on which the following holds:\\[1em]
For arbitrary subgraphs $E_b \subset E$ satisfying that for any vertex $i$, the number of edges emanating from $i$ which are corrupted is bounded by $\gamma n$ and for arbitrary pairwise direction corruptions $\mathbf{v}_{ij} \in \mathbb{S}^2$ for $(i,j) \in E_b$,  the convex program \eqref{eq:ShapeFit} has a unique minimizer equal to $\left \{\alpha \Bigl(\mathbf{t}^{(0)}_i - \bar{\mathbf{t}}^{(0)} \Bigr)\right\}_{i \in [n]}$ for some positive $\alpha$ and for $\bar{\mathbf{t}}^{(0)} = \frac{1}{n}\sum_{i\in [n]} \mathbf{t}^{(0)}$. 
\end{thm}
Both of these theorems follow from the much more general Theorem 3 in~\cite{ShapeFit} which establishes that ShapeFit succeeds in recovering locations from corrupted relative directions under broad deterministic assumptions on the measurement graph $G$ and the geometry of locations. Note that the setting for theorems about ShapeFit is that of robust statistics, in that the corruptions are assumed to be adversarial, and thus for a fixed set of locations and observations the theorem above establishes that ShapeFit succeeds uniformly in the data corruptions, provided there is a bound on the number of corrupted edges emanating from any particular location node.

The proof strategy for theorems about ShapeFit is a direct analysis of the optimality conditions of the convex objective function, which relies on the combinatorial propagation of various local geometric properties. For instance such a property is that if a collection of triangles share the same base and the locations opposite the base are sufficiently ``well-distributed'', then an infinitesimal rotation of the base induces infinitesimal rotations in edges of many of the triangles. It is then ensured that for each corrupted edge $(k,l)$ one can find sufficiently many triangles in the observation graph with two good edges and base $(k, l)$, with locations at the opposing vertices being ``well-distributed''. The full proof is however more nuanced and requires strongly using the constraints of the ShapeFit program. The other important local property is that for a tetrahedron in $\R^3$ with well distributed vertices, any discordant parallel deviations on two of its disjoint edges induce enough infinitesimal rotational motion on some other edge of the tetrahedron. Combinatorial propagation is then handled separately in two regimes of the relative balance of parallel deviations on the good and corrupted subgraphs.

The authors of \cite{ShapeFit} also provide empirical evidence supporting the main theoretical results about ShapeFit. These results demonstrate that the ShapeFit method can tolerate a larger fraction of pure outliers as compared to the LUD estimator for exact location recovery from randomly generated data, whereas in a mixed noise setting the LUD solver has a smoother change in its recovery performance (as corruption probability varies) and lower (higher) recovery errors in the high (low) corruption probability regime. More importantly, ShapeFit is empirically tested on a dozen standard photo-tourism datasets in \cite{ShapeKick}. A novel numerical framework for solving programs like ShapeFit and LUD is also introduced in \cite{ShapeKick}, which leads to an ADMM-based approach called ShapeKick which solves location recovery problems orders of magnitude faster than competing methods. A comparative study, with respect to competing methods such as LUD and 1DSfM, demonstrating the accuracy and high efficiency of ShapeFit and ShapeKick is also provided in \cite{ShapeKick}.

\subsection{Other Methods}
\begin{figure}[!htbp]
\begin{center}
\includegraphics[trim=0cm 0cm 0cm 0cm, clip=true, width=0.95\linewidth]{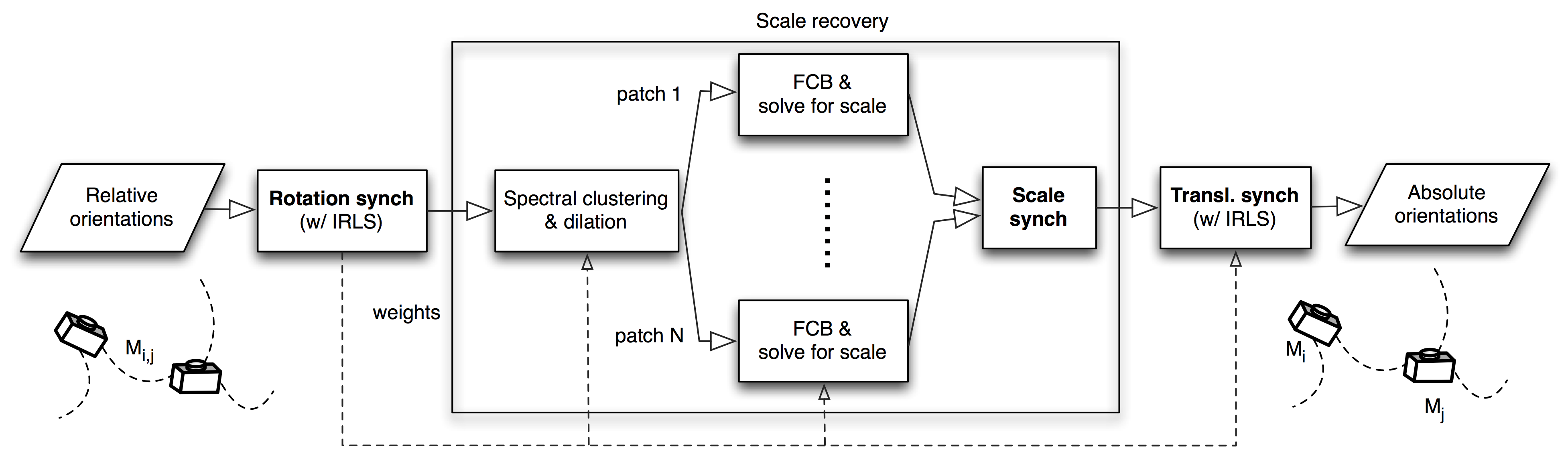}
\end{center}
\caption{A depiction of the group synchronization pipeline (GSP) in~{\rm\cite{ArrigoniSyncSfM}}, including the group synchronization stages in bold (figure courtesy of~{\rm\cite{ArrigoniSyncSfM}}).\label{fig:GSPdiag}}
\end{figure}

One of the methods deviating from the generic recipe considered above is studied in~\cite{GovinduLieAlg}. In this method, camera orientations and locations are jointly estimated using an \textit{iterative averaging method on the Lie algebra} of the group of Euclidean motions. Although~\cite{GovinduLieAlg} lacks theoretical and strong empirical stability guarantees, its approach is an efficient alternative for joint motion estimation in the low noise regime. Another recent method slightly deviating from the generic recipe is introduced in~\cite{ArrigoniSyncSfM}. Termed as the \textit{group synchronization pipeline (GSP)}, this method is based on a three step estimation, via (robust) synchronization, of the camera orientations in $\mbox{SO}(3)$, baseline scales\footnote{We note that the synchronization of the baseline scales is performed via a distributed, two-step method in~\cite{ArrigoniSyncSfM} in order to mitigate error propagation, where each step uses spectral methods to produce its estimates.} (i.e., $\alpha_{ij}$ in (\ref{eq:AltLinSys})) in $\R^+$, and the camera locations in $\R^3$ (cf.~Figure~\ref{fig:GSPdiag}, courtesy of~\cite{ArrigoniSyncSfM}, for a depiction of the GSP). 
The term \textit{synchronization} here refers to a class of methods aiming to (globally) estimate a set of mathematical group elements $\{g_i\}$ from a potentially noisy and incomplete set of observations of their pairwise ratios\footnote{Note that such an estimate can be obtained only up to a global multiplication $gg_i$ of the group elements by a group element $g$, since the ratios $g_i^{-1}g_j$ are invariant to such transformations.} $g_i^{-1}g_j$. The term was originally coined in~\cite{AmitAngSync}, which studies the problem of estimating planar rotations from measurements of their pairwise ratios. As an instance of a synchronization problem, in order to estimate the camera orientations $R_i\in\mbox{SO}(3)$ (represented as $3\times3$ orthonormal matrices with determinant one) from measurements $\hat{R}_{ij}$ of their ratios $R_{ij} = R_i^TR_j$, we can consider the problem
\begin{equation}
\label{eq:SYNC}
\begin{aligned}
&\underset{{\scriptstyle \{R_i\}}}{\text{minimize}}
& & \sum_{i\sim j}  \left\| R_i-R_j\hat{R}_{ij}^T\right\|_F^2 \\
& \text{subject to}
& & \{R_i\}\subseteq\mbox{SO}(3) \ ,
\end{aligned}
\end{equation}
which is a non-convex, difficult problem due to the set of constraints. However, the problem (\ref{eq:SYNC}) (and its variations using robust cost functions instead of the sum of squared deviations in (\ref{eq:SYNC})) was shown in the literature to admit high quality approximations via convex relaxations and spectral methods (cf., e.g., \cite{MicaAmitSfM,CvXSfM,GovinduRot,LanhuiSync,Mihai3DASAP}). The estimates of the first two steps in~\cite{ArrigoniSyncSfM} are fused into the available data in the last step to produce the final location estimates via an iteratively reweighted least squares (IRLS) solver for the resulting linear system of equations. The high quality of the overall approach in~\cite{ArrigoniSyncSfM} is empirically presented on large datasets. Instead of solely relying on relations between pairs of cameras, \cite{JiangTriLin} introduces a linear method based on minimizing the error in a set of geometrical relations among triplets of cameras (for details, cf. \S4.1 in~\cite{JiangTriLin}), which requires the knowledge of \textit{ratios} of baseline lengths computed from locally reconstructed scene points. Additionally, to improve the quality of the location estimates in the presence of outliers among pairwise measurements, \cite{JiangTriLin} employs various heuristic outlier rejection techniques. We note that, although this linear method has some desirable properties like being applicable to collinear motion, computational efficiency, etc., it is rather sensitive to outlier measurements, and hence the outlier rejection techniques proposed by~\cite{JiangTriLin} are of critical importance to maintain stability. Another alternative location estimation approach introduced in~\cite{MultiLinear}, which aims to tame the instability resulting from unknown scales, uses two-view reconstructions for camera pairs. The main idea is to obtain relative scales and translation between the two-view reconstructions sharing sufficiently many $3$D points, and to use this information in the camera location and scale estimation. The pipeline in~\cite{MultiLinear} also produces an initial $3$D structure estimate, which, together with its motion estimate, can be used to initialize reprojection error minimization algorithms for final structural refinement.

\section{Structure Estimation}
\label{sec:StructEst}
This section considers some of the key contributions in the literature concerning mainly the $3$D structure estimation part of the SfM problem. These works have a relatively wide spectrum including incremental and global methods, joint structure and motion estimation methods, methods specifically targeting very large scale SfM instances, etc. Typically, the number of $3$D structure points in an SfM instance is much greater than the number of corresponding cameras (e.g., for an unordered collection of about $10^3$ cameras, it is reasonable to expect on the order of $10^6$ structure points for a rich $3$D structure reconstruction). Therefore, the fundamental challenge for a structure estimation method is to \textit{efficiently} process the given data related to the estimation of $3$D points in a consistent way, while maintaining robustness to noise in the data.  

The primary class of algorithms for joint refinement of the $3$D structure, the camera motion and, possibly, also the intrinsic camera parameters (also termed as the camera calibration parameters) is known as \textit{bundle adjustment methods}. As indicated in the excellent survey~\cite{BundleAdjustment} of bundle adjustment (cf. Figure~9 in~\cite{BundleAdjustment}), the origin of bundle adjustment can be traced back to the works of Gauss and Legendre on least squares estimation in the late eighteenth and the early nineteenth centuries. However, in this section, we will be focusing on the more recent developments on bundle adjustment, and more generally on structure estimation methods that mainly aim to facilitate the processing of large, unordered sets of images.

In~\cite{BundleAdjustment}, several different aspects of bundle adjustment, including the basic projection model and problem parametrization, error modeling and the choice of cost function, main types of numerical optimization algorithms, the sparse problem structure as induced by the network of variables (and how to exploit it to improve reconstruction accuracy), gauge invariance, and quality control to detect outliers and characterize the overall accuracy of the estimated parameters, are discussed. The projection model and the problem parametrization are presented in a generalized, rather abstract framework to incorporate various scene models (e.g., the relatively simpler isolated $3$D features made up of points, lines, planes, etc.~or more complicated models involving dynamics, photometry, complex objects linked by constraints, etc.) and camera models (e.g., perspective, affine or orthographic projections, or less frequently encountered models such as rational polynomial cameras). To define a \textit{feature prediction error} (also sometimes referred to as the \textit{total reprojection error}), \cite{BundleAdjustment} considers a simple scene model comprising static, individual (abtract) $3$D features $\mathbf{X}_j$ imaged by a set of cameras corresponding to the pose and intrinsic calibration parameters $\mathbf{C}_i$. In this framework, we are given with noisy measurements $\bar{\mathbf{x}}_{ij}$ of the image features $\mathbf{x}_{ij}$ representing the true image of features $\mathbf{X}_j$ in image $i$, that are assumed to satisfy a \textit{predictive model} given by $\mathbf{x}_{ij} = \mathbf{x}(\mathbf{C}_i,\mathbf{X}_j)$. The feature prediction error for the measurements $\bar{\mathbf{x}}_{ij}$ are then defined by
\begin{equation}
\label{eq:BAerror}
\Delta\mathbf{x}_{ij}(\mathbf{C}_i,\mathbf{X}_j) \defeq \bar{\mathbf{x}}_{ij} - \mathbf{x}(\mathbf{C}_i,\mathbf{X}_j)
\end{equation}
To reconstruct the scene, a measure of these prediction errors are minimized, where bundle adjustment corresponds to the \textit{refinement} part of this optimization, and requires an initial set of estimates for the camera and structure parameters (e.g., computed using a simpler technique). In summary, as noted in~\cite{BundleAdjustment}, ``bundle adjustment is essentially a matter of optimizing a complicated nonlinear cost function (the total prediction error) over a large nonlinear parameter space (the scene and camera parameters)'', the accuracy of which is highly dependent on the initial estimates of camera and structure parameters (explaining, e.g., the wide variety of initial motion estimation methods), and on the choice of the parametrization of the parameter space (cf. \S2.2 in~\cite{BundleAdjustment} for a more detailed discussion). Additionally, \cite{BundleAdjustment} studies the choice of the cost function defining the measure of the errors in (\ref{eq:BAerror}), reaching to the main conclusion that robust, statistically-based error metrics allowing the presence of outliers in the measurements are the proper choices. Three main categories of algorithms for the numerical optimization of the chosen cost function are also discussed in detail in~\cite{BundleAdjustment}, namely the second-order Newton-style methods, first order methods (demonstrating linear asymptotic convergence) and the sequential methods incorporating a series of observations one-by-one (as opposed to solving for the whole system globally). For the second order methods, which have fast asymptotic convergence rates but relatively high costs per iteration, several different techniques for improving efficiency, mainly exploiting the sparsity patterns of the Hessian of the cost function, are explored in~\cite{BundleAdjustment}. Also, various first order methods, including the simple gradient descent, linear and nonlinear Gauss-Seidel methods, Krylov subspace methods, are discussed. As for the third category of sequential algorithms, low-rank update techniques for updating the Hessian and various filtering techniques are considered. \cite{BundleAdjustment} also provides methods for handling theoretical and algorithmic difficulties that arise from several gauge symmetries involved in the problem, and (essentially) concludes with various analytical and heuristic techniques for quality control, that are used to maintain algorithmic stability and robustness to outlier measurements, and to characterize the overall accuracy of the estimates.

\begin{figure}[!htbp]
\begin{center}
\includegraphics[trim=0cm 0cm 0cm 0cm, clip=true, width=0.6\linewidth]{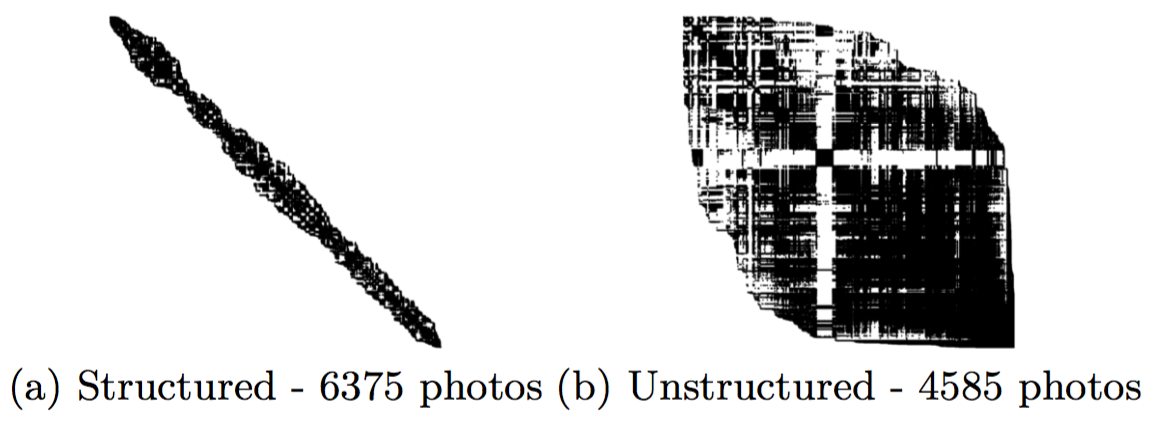}
\end{center}\vspace{-0.05in}
\caption{Sparsity patterns of the adjacency matrices of the connectivity graphs for two sets of images: {\rm(a)} A structured set of photos (taken from a moving truck) {\rm(b)} An unstructured collection of community photos (corresponding to a search of the term ``Dubrovnik'' in Flickr). Figure courtesy of~{\rm\cite{SnavelyBALarge}}.\label{fig:SnavelyBALarge}}
\end{figure}

A relatively recent work on large scale bundle adjustment (for tens of thousands of images) is provided in~\cite{SnavelyBALarge}. The main challenge identified in~\cite{SnavelyBALarge} is the low sparsity levels (i.e., relatively higher denseness) present in the connectivity graphs\footnote{Here, ``connectivity graph'' refers to the graph, the nodes of which denote the cameras corresponding to the images, and the edges of which exist between camera pairs having sufficiently many corresponding feature points in their corresponding images.} of \textit{unstructured} sets of images, such as community photo collections, as opposed to the higher sparsity of the \textit{structured} image sets (e.g., street-level datasets, such as images captured from a moving truck), cf.~Figure~\ref{fig:SnavelyBALarge} for an example (figure courtesy of~\cite{SnavelyBALarge}).
The decrease in sparsity increases the computational cost of sparse factorization methods employed in classical bundle adjustment methods drastically. To address this difficulty, the inexact (truncated) Newton type algorithm introduced in~\cite{SnavelyBALarge} makes use of conjugate gradients for the Newton step computation combined with various preconditioners. In more detail, \cite{SnavelyBALarge} considers a nonlinear least squares\footnote{As noted in~\cite{SnavelyBALarge}, the simple nonlinear least squares setup encapsulates cost functions more general than the $\ell_2$ norm, e.g. robust cost functions like Huber norm can be used by casting the problem as a reweighted nonlinear least squares instance.} bundle adjustment problem stated (abstractly) as
\begin{equation}
\label{eq:BASnavely}
\underset{{\scriptstyle \mathbf{x}}}{\text{minimize}} \ \ \frac{1}{2}\|\mathbf{F}(\mathbf{x})\|^2 \ ,
\end{equation}
where $\mathbf{x}\in\R^p$, $\mathbf{F}:\R^p\rightarrow\R^q$. Let $J(\mathbf{x})\in\R^{q\times p}$ denote the Jacobian of $\mathbf{F}$ at $\mathbf{x}$ and let $\mathbf{g}(\mathbf{x}) \defeq \nabla\frac{1}{2}\|\mathbf{F}(\mathbf{x})\|^2 = J(\mathbf{x})^T\mathbf{F}(\mathbf{x})$. \cite{SnavelyBALarge} considers a Levenberg-Marquardt (LM) approach to solve (\ref{eq:BASnavely}), which, to compute the update $\mathbf{x}\rightarrow\mathbf{x}+\Delta\mathbf{x}$ at each iteration (while ensuring convergence by controlling the step size $\|\Delta\mathbf{x}\|$), solves the regularized least squares problem
\begin{equation}
\label{eq:BASnavelyLMUpdate}
\underset{{\scriptstyle \Delta\mathbf{x}}}{\text{minimize}} \ \ \frac{1}{2}\left(\|J(\mathbf{x})\Delta\mathbf{x} + \mathbf{F}(\mathbf{x})\|^2 + \mu \|D(\mathbf{x})\Delta\mathbf{x}\|^2\right) ,
\end{equation}
where $D(\mathbf{x})$ is a nonnegative diagonal matrix (e.g., given by $D(\mathbf{x})_{ii} = \sqrt{\sum_{j}J(\mathbf{x})_{ji}^2}$) and $\mu > 0$ determines the regularization strength. Note that, if the regularized Hessian $H_{\mu}(\mathbf{x})\defeq J(\mathbf{x})^TJ(\mathbf{x}) + \mu D(\mathbf{x})^TD(\mathbf{x})$ is positive definite, we can solve (\ref{eq:BASnavelyLMUpdate}) via solving the \textit{normal equations}
\begin{equation}
\label{eq:BANormalEqs}
H_{\mu}(\mathbf{x})\Delta\mathbf{x} = -\mathbf{g}(\mathbf{x}) \ .
\end{equation}
Here, the crucial point is that, $H_{\mu}(\mathbf{x})$ (usually) turns out to have a special block structure induced by the sparsity pattern of the bundle adjustment problem, which simplifies the solution to (\ref{eq:BANormalEqs}). For instance, assuming that the components of $\mathbf{F}(.)$ only depend on a single camera and/or a single $3$D feature (which is usually the case), then we get 
\begin{equation}
\label{eq:BAHessBlock}
H_{\mu} = \left[\begin{matrix} B & E \\ E^T & C \end{matrix}\right] \ ,
\end{equation}
where $B$ and $C$ are block diagonal (with blocks of size $<10$, and hence can be efficiently inverted), and $E$ is block sparse. Therefore, by also considering the appropriate block representations $\Delta\mathbf{x} = [\Delta\mathbf{y}, \Delta\mathbf{z}]$ and $-g = [v,w]$, and solving for $\Delta\mathbf{z}$ by $\Delta\mathbf{z} = C^{-1}(w-E^T\Delta\mathbf{y})$, we obtain
\begin{equation}
\label{eq:BAreducedsys}
\left[B-EC^{-1}E^T\right]\Delta\mathbf{y} = v - EC^{-1}w \ .
\end{equation}
Here the Schur complement $S$ of $C$ in $H_{\mu}$ given by $S = B-EC^{-1}E^T$, and known as the \textit{reduced camera matrix}, has a block sparse structure with a nonzero $ij$'th block if and only if the $i$'th and the $j$'th images have at least one common feature point. As noted in~\cite{SnavelyBALarge}, computationally speaking, the reduction of (\ref{eq:BANormalEqs}) to (\ref{eq:BAreducedsys}), via the \textit{Schur complement trick} given above, allows us to obtain a solution to (\ref{eq:BANormalEqs}) using classical Cholesky factorization methods (that do not exploit the sparsity pattern in $S$) in $\mathcal{O}(n^2)$ space and $\mathcal{O}(n^3)$ time complexity, for $n$ denoting the number of cameras. These computational costs become prohibitive for large datasets, and hence an alternative is to use decomposition algorithms that exploit the sparsity in $S$. However, as the datasets grow and the level and simplicity of sparsity in $S$ decreases (as argued by~\cite{SnavelyBALarge} for the case of community photo collections), even the construction and storage of $S$ becomes computationally challenging. As a result, instead of trying to solve (\ref{eq:BASnavelyLMUpdate}) exactly using, e.g., the Schur complement trick discussed above, \cite{SnavelyBALarge} proposes to use an inexact Newton method based on an approximate iterative linear system solver, like the conjugate gradients (CG), possessing a termination rule given by
\begin{equation}
\label{eq:BATermRule}
\|H_{\mu}(\mathbf{x})\Delta\mathbf{x}  + \mathbf{g}(\mathbf{x})\|\leq\eta_k\|\mathbf{g}(\mathbf{x})\| \ ,
\end{equation}
where $0<\eta_k\leq\eta_0<1$, for $k$ denoting the LM iteration number, is used as a forcing sequence to guarantee convergence. Additionally, in order to improve the conditioning of the linear system $H_{\mu}(\mathbf{x})\Delta\mathbf{x} = -\mathbf{g}(\mathbf{x})$ for higher CG convergence rates, \cite{SnavelyBALarge}, as its main contribution, considers several \textit{preconditioners} $M$ to rewrite the linear system as $M^{-1}H_{\mu}(\mathbf{x})\Delta\mathbf{x} = -M^{-1}\mathbf{g}(\mathbf{x})$ in order to improve the rate of convergence by minimizing the condition number of the matrix $M^{-1}H_{\mu}(\mathbf{x})$ (which directly determines the rate of convergence). \cite{SnavelyBALarge} concludes with extensive experimental stability and efficiency comparisons for their preconditioned CG method, with various preconditioners, and the classical methods factoring the Schur complement, obtaining accuracy and significant efficiency (both in time and in memory) improvements using their proposed algorithm.

For large, unordered, and hence computationally challenging datasets, an alternative global approach is introduced in~\cite{SnavelyDISCO}, where the formulation is based on first computing a coarse initial solution via a hybrid \textit{discrete-continuous optimization}, using discrete belief propagation (BP) within a Markov random field (MRF) formulation of pairwise camera and camera-point constraints to estimate camera parameters and a continuous LM improvement, and then refining the resulting solution with classical bundle adjustment. This method also employs various alternative sources of data like noisy geotags and vanishing point estimates extracted from images. More concretely, the method first solves for camera orientations by minimizing a robust cost function of the error in the (classical) pairwise rotational consistency equations (i.e., the equations $R_{ij} = R_i^TR_j$, for which we have access to noisy measurements of some of the relative rotations $R_{ij}$), using a mixture of BP (on a discretization of the $3$D sphere) and LM refinement. Having solved for the orientations $R_i$, the joint estimation of the camera locations and structure points is based on the minimization of (a robust cost function of) the total \textit{angles of deviation} between pairwise translations $\mathbf{t}_i-\mathbf{t}_j$ and pairwise directions $\mathbf{\gamma}_{ij}$ (cf., e.g., (\ref{eq:DirHomEq})) and between $3$D point-to-camera displacements (given by $\mathbf{P}_k-\mathbf{t}_i$ between the $k$'th $3$D point $\mathbf{P}_k$ and camera $i$) and ``ray directions'' (given in the noiseless case by $\mathbf{r}_{ik} = R_i^TK_i^{-1}\mathbf{p}_{ik}$, for the known intrinsic matrix $K_i$ of camera $i$, and the projective measurement $\mathbf{p}_{ik}$ of $\mathbf{P}_k$ by camera $i$). This minimization is again performed by a hybrid BP and LM method. Even though, in principal, such non-convex methods are highly sensitive to initial points, and hence susceptible to local minima, \cite{SnavelyDISCO} is able to improve its sensitivity and obtain, as empirically demonstrated, high quality reconstructions by also making use of external, additional sources of information like geotags and vanishing points. Additionally, the parallelizable structure of the methods in~\cite{SnavelyDISCO} allows the realization of significant improvements in computational efficiency.

\begin{figure}[!htbp]
\begin{center}
\includegraphics[trim=0cm 0cm 0cm 0cm, clip=true, width=0.95\linewidth]{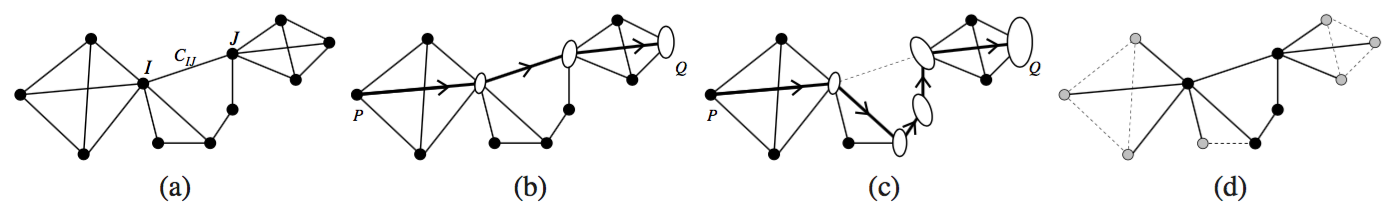}
\end{center}\vspace{-0.05in}
\caption{A toy example for the construction of the skeletal graph in~{\rm\cite{SnavelySkeletal}}: {\rm(a)} Nodes of the image graph represent images and its edges represent pairwise reconstructions (in fact, the image graph is directed and each connected pair has two directed edges, only one is shown for simplicity). Each (directed) edge $(I,J)$ is endowed with a covariance matrix $C_{IJ}$ representing the uncertainty in image $J$ relative to $I$ {\rm(b)} Relative uncertainty between the nodes $P$ and $Q$ is quantified by the trace of the covariance matrices (represented by ellipses) chained up along the shortest path (denoted using arrows) between the nodes {\rm(c)} Removing an edge lengthens the shortest path, resulting in larger estimated covariance (denoted by larger ellipses) {\rm(d)} The solid edges are those of the skeletal graph, while the dotted edges have been removed. The black (interior) nodes form the skeletal set, and are to be reconstructed first, while the gray (leaf) nodes are added to it using pose estimation. Skeletal graph is constructed by trying to minimize the number of interior nodes, while bounding the increase in relative uncertainty between all pairs of nodes in the original graph in {\rm(a)}. Figure courtesy of~{\rm\cite{SnavelySkeletal}}.\label{fig:SnavelySkeletal}}
\end{figure}

A relatively popular main approach to handle challenging datasets is to employ \textit{sequential processing} of the available images, that is to incorporate the images one by one, or in small groups, to obtain a solution for the whole set based on a ``core solution'' computed from a small subset of the available data. A systematic method of this sequential type was introduced in~\cite{SnavelySkeletal}, the main idea of which is to identify a small subset of the available images, named the \textit{skeletal set}, by first estimating the accuracy of pairwise reconstructions of overlapping images and then extracting a subset whose reconstruction accuracy approximates that of the total set. The remaining images are then incorporated by pose estimation and a final bundle adjustment can be used to improve accuracy. The computation of the skeletal set is rather complicated\footnote{For brevity, we skip the finer details of the computation of the skeletal set in~\cite{SnavelySkeletal}, such as removal of outlier pairwise reconstructions, removal of (duplicate) images with similar content, identification of \textit{infeasible paths} between triplets of images using higher-level connectivity structures named the \textit{pair graph} and its augmentation with the \textit{image graph}, etc.}, however the main idea is to identify a (directed) measure of proximity between cameras, identified as nodes of the \textit{image graph}, given by the (translational) uncertainty in the pairwise reconstruction (obtained by fixing the location and orientation of one of the cameras in the pair, and then repeating the same for the other) as quantified by the trace of the \textit{covariance} matrix given by the translational part of the inverse of the Hessian of the reduced camera system (\ref{eq:BAreducedsys}) (i.e., the Schur complement $S = B-EC^{-1}E^T$ for the camera pair), cf.~Figure~\ref{fig:SnavelySkeletal} for a simplified depiction (image courtesy of~\cite{SnavelySkeletal}).
The skeletal set problem is to identify the smallest subset of cameras, subject to the constraint that the distance (length of the shortest path, as defined by the total uncertainty of the path) between any pair of cameras in the (directed) subgraph induced by this subset is at most $t$ times longer than their distance in the original digraph (induced by the set of all cameras). The factor $t$ is named as the \textit{stretch factor}, and controls the amount of additional uncertainty introduced between camera pairs by removing cameras from the available dataset. Also, the skeletal set is required to satisfy the constraint of being uniquely realizable, which is modeled and controlled via a structure termed as the \textit{pair graph} in~\cite{SnavelySkeletal}. Unfortunately, this problem (even if we ignore the additional realizability constraint), known as \textit{computing a minimum $t$-spanner} for the digraph of camera pairs, is NP-complete for general graphs. Hence, \cite{SnavelySkeletal} employs a two-step heuristic approximation to compute the skeletal set (cf.~\S4 in~\cite{SnavelySkeletal} for the details). The empirical results provided in~\cite{SnavelySkeletal} for their method based on the skeletal sets demonstrate the significant increase in the overall SfM efficiency, and the accuracy of the reconstructed structure as compared to applying existing SfM techniques on the full set of images.

Another incremental method, similar in spirit to~\cite{SnavelySkeletal}, is studied in~\cite{HavlenaGraphOptim}. The main idea in~\cite{HavlenaGraphOptim} to improve the efficiency of the overall SfM pipeline is to identify a small subset of input images by computing an approximate \textit{minimally connected dominating set}, for which each of the remaining images has significant overlaps with at least one of the images in this set, and to employ \textit{task prioritization} to promote easier instances of image matching. Additionally, instead of computing actual image matchings, which is computationally prohibitive for thousands of cameras if performed on each pair of images, \cite{HavlenaGraphOptim} employs fast image indexing techniques based on large image vocabularies to measure visual overlaps between images. To identify image similarities, \cite{HavlenaGraphOptim} uses a ``bag-of-words approach'' (cf.~\S2.1 in~\cite{HavlenaGraphOptim} for the details). The computation of a minimally connected dominating set for the induced (unweighted, undirected) graph, defined as a minimum sized connected subgraph, of which each vertex of the original graph is either a member or a neighbor, is in general NP-hard. Hence, \cite{HavlenaGraphOptim} uses a polynomial time approximation algorithm from the literature for this task (cf. Algorithm 1 in~\cite{HavlenaGraphOptim}). To construct the $3$D structure, \cite{HavlenaGraphOptim} makes use of the \textit{atomic $3$D models from image triplets} approach of~\cite{HavlenaGroups} and grows the structure by using a prioritization queue, which orders different tasks of either creating a new atomic $3$D model or incorporating one image to a given $3$D model (cf.~\S3 in~\cite{HavlenaGraphOptim}). The authors conclude with various experimental results, including results on image sets captured by omnidirectional cameras, demonstrating the efficiency and accuracy of their approach.

A notable work of an incremental nature is the \textit{Photo Tourism} system of~\cite{SnavelyData}, that was also discussed in the introduction. Although the SfM part of~\cite{SnavelyData} is primarily based on a rather classical, generic type of a sequential incorporation of images one-by-one using the sparse bundle adjustment algorithm of~\cite{SBA}, it provides a very rich set of tools for content exploration, such as $3$D scene visualization, object-based photo browsing, geo-location from images, object annotation, etc.~(cf.~Figure{fig:SnavelyPhotoTourism}, courtesy of~\cite{SnavelyData}, for a simplified illustration of the Photo Tourism system). 
\begin{figure}[!htbp]
\begin{center}
\includegraphics[trim=0cm 0cm 0cm 0cm, clip=true, width=0.95\linewidth]{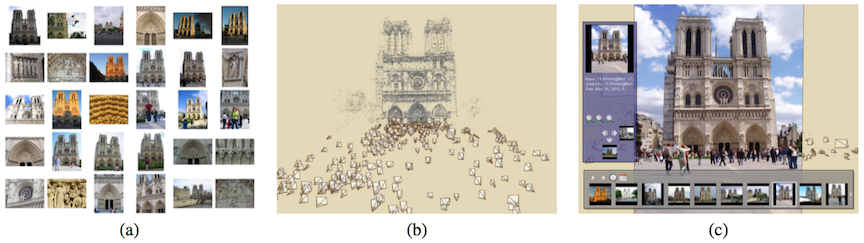}
\end{center}\vspace{-0.05in}
\caption{Photo tourism system of~{\rm\cite{SnavelyData}}: {\rm(a)} An unstructured collection of images, e.g.~downloaded from the Internet, as input to the system {\rm(b)} SfM step composed of camera motion estimation and 3D structure reconstruction {\rm(c)} Content exploration, such as 3D scene visualization, object-based photo browsing, geo-location from images, object annotation, etc. Figure courtesy of~{\rm\cite{SnavelyData}}.\label{fig:SnavelyPhotoTourism}}
\end{figure}
For the sequential SfM, \cite{SnavelyData} identifies an initial pair of cameras sharing a large set of feature points, which also have baseline to improve the accuracy of the reconstructed initial $3$D points. The selection criteria for a new camera is to have a maximal number of feature points that correspond to the already reconstructed $3$D points. This iterative procedure is repeated, until no remaining camera observes any reconstructed $3$D point (for additional details, such as detection of outlier tracks, geo-registration, specific algorithms used, etc., cf.\S4 in~\cite{SnavelyData}). Even though the overall computational cost of the SfM pipeline proposed in~\cite{SnavelyData} can be quite high for relatively large datasets (e.g, processing and matching of the $2,635$ images in the Notre-Dame dataset, and the registration of $597$ of them is reported to take \textit{two weeks}), it provides an accurate alternative for SfM, and more importantly, the rich set of tools it provides for content exploration strikingly demonstrates the wide range of applicability and usefulness of SfM.

Yet another remarkable SfM approach for processing large (specifically, \textit{city scale}), unordered datasets is introduced in~\cite{SnavelyRome}. The key paradigm explored in~\cite{SnavelyRome} is the \textit{parallel processing} of the various stages in SfM, and hence its main contributions are the identification of challenges and introduction of solutions in the parallelization of the SfM pipeline. In its multi-stage image matching method, where each stage consists of a proposal (of possible image pairs sharing common features) step and a verification step, \cite{SnavelyRome} employs several different techniques, such as vocabulary tree based whole image similarity, query expansion, etc. (cf. \S2.2 and \S2.3 in~\cite{SnavelyRome} for details), while simultaneously controlling the computational efficiency of the distributed framework. For the SfM part, \cite{SnavelyRome} uses the skeletal set method of~\cite{SnavelySkeletal} to significantly reduce the computational cost involved, and the initial camera motion and $3$D structure for the skeletal set are computed via the sequential SfM method of~\cite{SnavelyData}. To improve upon the computed motion and structure, bundle adjustment is performed, where, depending on the size of the problem, \cite{SnavelyRome} chooses between a preconditioned CG method (as in~\cite{SnavelyBALarge}) for large datasets, and an exact step LM algorithm for smaller datasets. Finally, the accuracy and efficiency of the overall approach in~\cite{SnavelyRome} is demonstrated through extensive experimental results for various city scale datasets.

Interestingly, after camera rotations have been estimated, the rest of the SfM pipeline may be reformulated as a bipartite location recovery problem. Namely, camera locations and 3D structure points form the two sets of nodes of the bipartite graph, and directions between cameras and scene points obtained from correspondences between pairs of views form the edges. This observation, coupled with fast algorithms for ShapeFit and LUD, has potentially substantial implications for real-time robotics applications, as discussed in \S\ref{sec:SLAM}. In follow up work to \cite{ShapeFit}, the same authors prove a set of corruption-robust exact recovery theorems for ShapeFit in the bipartite setting \cite{ShapeFitJoint}. As obtained after the correspondence estimation step of the SfM pipeline, \cite{ShapeFitJoint} assumes the availability of measurements (partially corrupted with adversarial noise) of camera-to-point directions $(\mathbf{t}_i-\mathbf{p}_j)/\|\mathbf{t}_i-\mathbf{p}_j\|$, for $\mathbf{t}_i\in\R^d$ denoting the $i$'th location and $\mathbf{p}_j\in\R^d$ denoting the $j$'th structure point. By using the convex ShapeFit algorithm introduced in~\cite{ShapeFit} for the simultaneous estimation of locations and structure points, \cite{ShapeFitJoint} proves a (high-dimensional) exact recovery result (cf. Theorem 1 in~\cite{ShapeFitJoint} for the precise statement of this main result) for a set of i.i.d.~Gaussian locations and structure points with high probability if the underlying bipartite measurement graph is sampled from an Erd\"{o}s-R\'{e}nyi ensemble, the problem dimension $d$ is sufficiently large, and at most a constant fraction of direction measurements involving any particular location or structure point are corrupted. Although this result is in the high-dimensional regime, it is the first \textit{simultaneous} exact location and structure recovery result for partially corrupted camera-to-point direction measurements. The proofs in \cite{ShapeFit} crucially rely on the existence of sufficiently many triangles in random graphs, while any bipartite graph cannot contain a triangle. The results in \cite{ShapeFitJoint} establish the necessary properties for exact recovery in high dimensions from corrupted relative-direction observations, similar to Theorem 1 in \cite{ShapeFit}, by relying instead on the existence of randomly distributed four-cycles in random bipartite graphs. \cite{ShapeFitJoint} also provides empirical evidence on simulated data supporting its theoretical findings, and \cite{ShapeKick} evaluates bipartite ShapeFit on real datasets, providing evidence of robust recovery in the bipartite setting. 

\section{Simultaneous Localization and Mapping (SLAM)}
\label{sec:SLAM}
Simultaneous localization and mapping (SLAM) consists of estimating a depth map of the $3$D environment, as well as localizing the agent which is moving throughout this environment. SLAM is a fundamental problem in robotics, providing crucial information for autonomous navigation of cars, drones and consumer robots. Pre-built maps of the environment can be used to provide much useful information for navigation, provided that localization within those maps can be performed accurately. The correct assertion that a vehicle or robotic entity is in a previously observed location is called a loop closure. SLAM is effectively a special case of the SfM problem, in that it is specific to robotics and augmented/virtual reality applications, yet retains many of the same difficulties. 

A multitude of combinations of sensing modalities have been used for SLAM, including LIDAR, inertial sensors, GPS, and vision. LIDAR is a form of active sensing, which consists of measuring laser time of flight to infer depth in real time. While LIDAR to a large extent trivializes the depth estimation problem and also eases the task of loop closure, it is an expensive sensor with low resolution and suffers performance degradation in suboptimal weather conditions such as rain and fog. Nonetheless, LIDAR is the centerpiece of many of today's approaches to autonomous navigation, by allowing pre-mapping of urban environments and accurate loop closures in such maps, which provides information about the static part of the environment in real time. Localization can also be aided by differential GPS, which has very high accuracy, but is also expensive. The holy grail of SLAM research is a purely vision based approach, since vision sensors are cheap and ubiquitous, and provides very rich and high resolution information about the environment. However, due to the aperture problem and various other ambiguities of the image formation process, using vision alone for SLAM is a difficult task and remains at the forefront of robotics research, in particular due to its high computational complexity. We will cover in this section state of the art vision based and visual-inertial SLAM approaches as well as some of the particular difficulties faced in robotics applications. 

\subsection{Novel Approaches to and Limitations of Visual SLAM}
As outlined above, the SfM pipeline (and similarly a SLAM pipeline that doesn't rely on pre-built maps) begins with 1) establishing feature correspondences and then 2) estimating camera orientations from relative orientations. Once camera orientations have been estimated, the task is to recover camera locations and $3$D structure points, which can be done in several ways. The standard pipeline involves first solving for camera locations from relative directions between cameras, obtained from epipolar geometry and orientation estimates. Several algorithms, as described in~\S\ref{sec:CamLocEst}, can be used for this step. Once camera locations have been recovered, $3$D structure may be obtained via triangulation. Often, the resulting camera pose and $3$D structure estimates are not as accurate as desired, and there is a global step which estimates these quantities simultaneously provided an initialization, which is called bundle adjustment (cf.~\S\ref{sec:StructEst}). Standard approaches to bundle adjustment are computationally prohibitive for real-time applications. Recent progress in algorithms and numerical methods allow for fast solutions of the location recovery problem. In particular a numerical framework based on ADMM and specially tailored for SfM has been developed in~\cite{ShapeKick}, which allows solving the LUD and ShapeFit programs $10$-$100$ times faster than competing approaches. This capability offers a new path toward real-time SLAM by re-formulating the SfM pipeline. Namely, after obtaining camera orientations, the point correspondences give relative directions between camera and $3$D structure point locations. Thus, we may frame estimating camera locations and $3$D structure points as a location recovery problem on a bipartite graph. ShapeFit and LUD can be applied to such a bipartite formulation, and ShapeFit has been proven to recover exactly locations and $3$D structure in this regime under suitable assumptions. Empirical results demonstrate the effectiveness of ShapeFit and LUD in the bipartite regime. There is potential to do away with bundle adjustment if the output of this pipeline is accurate enough, which would enable usage in real time robotics applications. The entire process depends on the accuracy of the estimated camera orientations and in practice, accelerometers are relatively accurate at providing such estimates.

The main difficulty in vision-based SfM systems is the abundance of outliers in correspondence estimates, which are due to the highly ambiguous nature of local photometric information due to illumination changes, occlusions, specularities, repetitive structures and objects that move independently of the scene. In offline applications such as photo-tourism~\cite{SnavelyData}, it is possible to lower the outlier rates in correspondences to still high but acceptable levels (say $40$ percent) by using RANSAC~\cite{ransac}, which is a probabilistic technique that in general randomly samples a set of observations, fits a model, and validates that model by how many of the other observations it approximately fits. In the case of SfM, RANSAC exploits the 8-point algorithm, which given a set of 8 accurately estimates correspondences between a pair of views, estimate the relative orientation and direction between camera locations. Having obtained a candidate relative orientation and translation between camera views, it is possible to validate correspondences. While effective, RANSAC is ultimately a combinatorial procedure, and thus it is only tractable to run to completion in offline applications such as photo-tourism. In real-time settings, tradeoffs have to be made with the use of RANSAC, at times leaving the fraction of outliers in correspondences as high as $80$ or $90$ percent. Such very high outlier ratios make SfM intractable via methods like ShapeFit and LUD, and extra information needs to be taken into account to enable efficient SfM in such regimes. 

\subsection{Visual-Inertial Navigation Systems}
To alleviate the issue outlined in the previous paragraph, it is common practice in applications to combine visual information with input from inertial sensors (accelerometers and gyrometers), to provide estimates of $3$D orientation and position of the sensor platform, also called visual-inertial navigation (VINS). Such an approach is prevalent in applications such as augmented and virtual reality. It is common to model the visual inertial navigation problem as the evolution of a dynamical system, the states of which signify $3$D pose of the sensor platform as an element of the Euclidean group of motions $\mbox{SE}(3)$, which produces sensor measurements as outputs up to some uncertainty. It is a natural first step to analyze the observability of such a system, under assumptions on model parameters such as calibration coefficients, the sensor biases, and a generative ``noisy'' input which drive the evolution of the system. While most models treat sensor bias as noise independent of other states, this assumption is frequently violated in practice. In~\cite{Tsotsos1}, it is shown that in the absence of such an independence assumption, the resulting model is not observable. The problem is then recast as one of sensitivity analysis, with results of~\cite{Tsotsos1} deriving an explicit characterization of the set of trajectories indistinguishable under the observed measurements, as a function of the properties of the sensor and sufficient excitation conditions (the properties of motion of the sensor platform), which can be used for analysis and validation purposes. 

In~\cite{Tsotsos2} the authors consider the VINS task while specifically focusing on the handling of outliers. This task falls under the umbrella of robust statistics and most VINS systems employ some for of outlier/inlier testing. In~\cite{Tsotsos2}, authors derive the optimal outlier discriminant, showing that it's intractable, motivating state augmentation with a delay-line in the model and examining various analytical and empirical approximations. Compared to decoupled systems, i.e.~those in which pose estimates are computed from visual and inertial measurements individually and then fused, the approach in~\cite{Tsotsos2} is guaranteed to stay within a bounded set of the state trajectories. 

\subsection{Application-specific Difficulties}
Structure from motion is a well developed subfield of computer vision, and has lead to a number of successful commercial applications. However, the distribution of camera locations strongly affects the performance of SfM and SLAM algorithms, and is dictated by the application at hand. For instance, photo-tourism has been successfully commercialized and is one of the easier cases of SfM because camera locations tend to be well-distributed in space, thus providing sufficiently diverse views of the scene which can be used to infer depth information. On the other hand, the application of autonomous driving requires handling scenarios in which the camera moves in essentially a straight line for long periods of time. There is a multitude of issues with this special case: 1) the most informative features (the ones that undergo the most motion) are in the periphery and quickly move out of the field of view, which explains the popularity of omnidirectional cameras for such applications, 2) the existence of many local minima in the least squares reprojection error, which is the basis for many algorithms, each corresponding to various ambiguities such as Necker-reversal, plane-translation and bas-relief and finally 3) Considering the worst case of moving exactly in parallel with the optical axis, it is evident that depth at the center pixel of the image is unrecoverable and this phenomenon induces many local minima around the true direction of translation. Semi-global approaches aim to overcome these issues, but they are too computationally intensive for real-time applications. One may ponder whether these ambiguities are inherent to the problem, or are a byproduct of the least squares reprojection functional. The work of \cite{VGSoattoMovingForward} shows that the answer is at least partly the latter, by showing that by enforcing a bound on the depth of the scene makes the reproduction error continuous. However, local minima are still present, as evidenced by empirical results. 

\subsection{Direct vs.~Feature-based Methods}
An inherent challenge of applications, such as robotics and augmented/virtual reality, is transitioning between differently scaled scenes, for instance from the indoors to the outdoors. Sensors which provide scale measurements, such as stereo or LIDAR, have a limited scale range in which they perform well. For instance, depth that far exceeds the baseline of a stereo rig is difficult to infer via stereo without sufficiently high image resolution, which in turn impacts realtime computational tractability. Meanwhile, monocular SLAM is immune to this issue since it is inherently scale-invariant.  At the same time, due to the fact that absolute scale cannot be inferred in monocular SLAM, scale estimates are subject to drift over time. 

SfM approaches fall into two categories: feature based and direct methods. Feature based approaches, as described above, decouple the problem into two steps: 1) features are extracted from images and 2) camera pose and $3$D scene structure is computed based on 1) only. Naturally, the efficacy of such approaches depends crucially on the admissible set of features, which in many approaches are limited to a sparse set of key-points in each image. In contrast, direct methods for visual odometry attempt to utilize all the photometric information in images by optimizing for camera pose and $3$D structure directly. While direct methods have an information theoretic advantage over feature based ones, they are typically more computationally intensive. A method recently proposed in~\cite{LSDSLAM}, coined Large-Scale Direct monocular SLAM (LSD-SLAM), is capable of building consistent large scale maps of a $3$D scene in real-time, by estimating semi-dense depth maps via filtering and direct image alignment. The map of the scene is represented as a collection of keyframes with $3$D similarity transformations as constraints between poses at keyframes. This representation allows scale changes to naturally be taken into account and corrects accumulated drift. 

For alternative methods and further discussion on SLAM, we refer the reader to other sources, e.g., \cite{SLAMSurvey1,SLAMSurvey2,SLAMSurvey3}, and the references therein.

\section{Other Topics}
\label{sec:Misc}
This section shortly discusses some of the additional topics in the SfM literature not considered in the previous sections. These topics include methods for feature extraction and matching, alternative core measurement types and camera models, methods aiming to handle symmetries and ambiguities in images, and widely utilized sources of data and SfM software.

\subsection{Feature Description and Matching}
\label{subsec:FeatureExtraction}
\begin{figure}[!htbp]
\begin{center}
\includegraphics[trim=0cm 0cm 0cm 0cm, clip=true, width=0.75\linewidth]{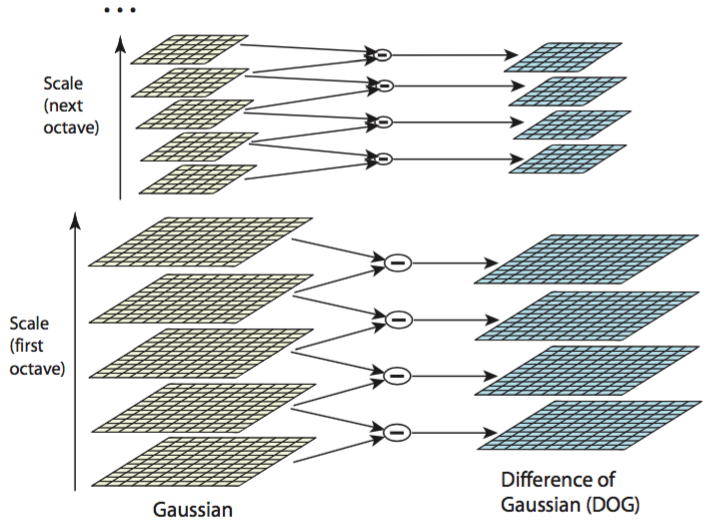}
\end{center}
\caption{The scale-space representation of an image in~{\rm\cite{SIFT}} using difference-of-Gaussians (DoG): For each octave of the scale space, the original image is first convolved with Gaussians to obtain the set of scale space images (on the left), and then adjacent Gaussian images are subtracted to produce the DoG images (on the right). After each octave, the Gaussian image is down-sampled by a factor of 2, and the process is repeated (figure courtesy of~{\rm\cite{SIFT}}).\label{fig:SIFTDoG}}
\end{figure}

\begin{figure}[!htbp]
\begin{center}
\includegraphics[trim=0cm 0cm 0cm 0cm, clip=true, width=0.85\linewidth]{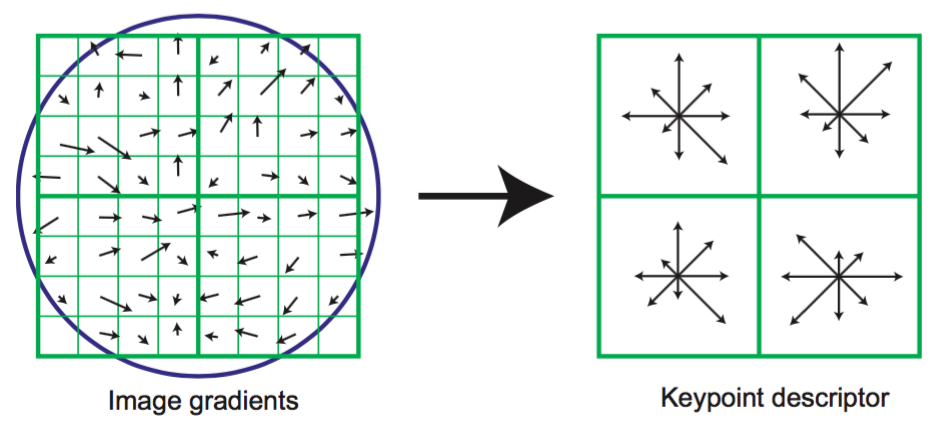}
\end{center}
\caption{To obtain a keypoint descriptor, first the gradient (magnitude and orientation) at each image sample point in a region around the keypoint location is computed (shown on the left) and then scaled by a Gaussian window (represented by the overlaid circle). These samples are then collected into orientation histograms representing the contents over $4\times4$ subregions (shown on the right), where the length of each arrow is given by the sum of the gradient magnitudes quantized to that direction within the region (for simplicity, the figure shows a $2\times2$ descriptor array computed from an $8\times8$ set of samples, whereas actually $4\times4$ descriptors computed from $16\times16$ sample arrays are used). Figure courtesy of~{\rm\cite{SIFT}}.\label{fig:SIFTDescriptor}}
\end{figure}

Although the processes of detection, description and matching of image features are usually considered to precede the SfM steps, here we touch upon some of the most frequently used methods for these tasks\footnote{The literature for the (classical) problem of \textit{detection} of local feature points (sometimes also referred to as feature extraction) is a vast accumulation of the works in roughly the last five decades, including techniques for detection of edges, corners, blobs and regions, segmentation based methods, machine learning based methods, which we do not intend to discuss here in detail. For an exposition of some of these methods, cf., e.g.,~\cite{FeatDetSurvey}.}. One of these methods is the \textit{scale invariant feature transform} (SIFT)~\cite{SIFT2,SIFT}, which has been extensively used as a building block of many of the existing corresponding point-based SfM methods (cf., e.g.,~\cite{SnavelyData,VisualSfM,SnavelyDISCO,CvXSfM,LUD,MicaAmitSfM}). In essence, SIFT transforms an input image into a collection of \textit{local features} (e.g., represented as vectors in some Euclidean space), which are invariant to scaling and rotation of the underlying image, and partially invariant to $3$D camera viewpoint and to changes in illumination. The detection and description stages of SIFT are mainly composed of four steps. In the first step, potential key points, which are invariant to scale and rotation, are determined via detection of extrema in a \textit{scale-space representation} of the image, using difference-of-Gaussians (DoG). The scale-space representation of an image, depicted in Figure~\ref{fig:SIFTDoG} (courtesy of~\cite{SIFT}), is a parametrized collection of smoothed images, where the single parameter corresponds to the size of a kernel used to filter out fine structures in the image. 
SIFT computes scale-space representations, using DoG kernels, repeatedly for down-sampled copies of the original image, and then identifies locally extremal points on these images by simply comparing the pixel values with the nearest neighbors in the current and the adjacent scales. After identifying these candidate points, i.e.~the extrema in the scale-space representation, the next step is to (approximately) compute location, scale, and ratio of principal curvatures (in order to identify and reject the candidates having low contrast or that are poorly localized on an edge, cf.\S4 in~\cite{SIFT} for details). The third step aims to achieve invariance to rotation in the descriptor by assigning an orientation (to be phased out in the representation of the descriptor) to each keypoint based on local image gradient directions. At this stage, SIFT has assigned a scale, an orientation and a location to each remaining keypoint in the image, and the next step involves the computation of a descriptor for (the local regions around) each keypoint, while maintaining invariance to changes in illumination and $3$D viewpoint, as much as possible. By operating on the image having the scale closest to that of a keypoint to fix a level of Gaussian blur, SIFT firstly samples gradient magnitudes and orientations around the keypoint location, and rotates the coordinates of the descriptor and the gradient orientations relative to the keypoint orientation (computed in the previous step) in order to achieve rotational invariance, cf.~Figure~\ref{fig:SIFTDescriptor} (courtesy of~\cite{SIFT}) for a depiction. 
The descriptor is given as a collection of orientation histograms (having $8$ bin values corresponding to a weighted sum of magnitudes of gradients, whose directions are close to one of $8$ quantized direction) of $4\times4$ subregions around the keypoint location. The resulting $128$ dimensional descriptor is then normalized to have unit norm to attain invariance to homogeneous changes in illumination. To also improve invariance to non-linear changes in illumination, the entries of this unit norm feature vector are thresholded (to compensate for large relative changes in gradient magnitudes), and are normalized to have unit norm once more. In the last stage of matching the extracted feature vectors (i.e., descriptors) between images, the Euclidean distance between feature vectors is used to compute nearest neighbors. Matched points are then identified to be those closer to their nearest neighbors than a fixed threshold.

There are several works in the literature building upon the SIFT method. One such method, named PCA-SIFT~\cite{PCASIFT}, is different from SIFT only in its computation of descriptors from local neighborhoods around keypoints. Similar to SIFT, for a given keypoint with assigned scale and orientation, PCA-SIFT extracts a $41\times41$ patch around the keypoint (in the image corresponding to the assigned scale) and rotates the patch using the assigned orientation to attain rotational invariance. Then, a $2\times39\times39=3042$ dimensional vector of concatenated vertical and horizontal gradients for the $41\times41$ patch centered at the keypoint is computed. As in SIFT, this vector is normalized to have unit length, to improve invariance to changes in illumination. A low dimensional representation of this unit norm vector is the desired descriptor. To compute this low dimensional representation, PCA-SIFT initially computes (offline) an \textit{eigenspace} spanned by the top $n$ eigenvectors (for, e.g., $n=20$) of the covariance matrix of a large number (taken to be $21,000$ in~\cite{PCASIFT}) of $3042$ dimensional vectors corresponding to patches extracted from a diverse collection of images. The descriptor for each keypoint is then computed to be the orthogonal projection of the $3042$ dimensional vector corresponding to its local patch onto the $n$ dimensional pre-computed eigenspace. The computed descriptors are matched between images via the same method in SIFT. In~\cite{PCASIFT}, PCA-SIFT is also empirically demonstrated to be resilient to various deformations and noise in the images (also see~\cite{GLOH} for a comparative study). 

An alternative variation of SIFT, coined \textit{gradient location and orientation histogram} (GLOH), is introduced in~\cite{GLOH}\footnote{The main focus of~\cite{GLOH} is actually the performance evaluation of various descriptors, based on local regions, with respect to various factors influencing their performance. Hence, for the details of some of the other methods for feature extraction, we refer the reader to~\cite{GLOH}.}. In essence, GLOH uses a similar approach to PCA-SIFT to compute descriptors. The difference is mainly in its representation of the locations of the gradients in the neighborhood of the keypoints. Instead of dividing the local neighborhood (in the image having the assigned scale) of the keypoint using a rectangular grid, GLOH uses \textit{logarithmic polar coordinates} with three bins in radial direction and eight bins in angular direction (without dividing the central bin in angular directions), resulting in a total of $17$ locations. Also using a quantization of the gradient orientations into $16$ bins, GLOH computes a histogram of $16\times17 = 272$ bins. Similar to PCA-SIFT, this histogram is represented in a $128$ dimensional eigenspace computed from a collection of $47,000$ image patches (via PCA of the corresponding covariance matrix). \cite{GLOH} also provides extensive empirical results demonstrating the performance of GLOH relative to nine different methods including SIFT and PCA-SIFT.

For details and performance evaluation of some of the other approaches in the literature (including the computationally efficient \textit{speeded up robust features} (SURF) method of~\cite{SURF}) for detection and description of local features, we refer the reader to~\cite{DescSurvey2}.

\subsection{Symmetries and Ambiguities in Images}
\label{subsec:Ambiguities}
A fundamental difficulty for SfM methods based on \textit{local features} in images is the presence of distinct objects in the $3$D structure that look similar. These similarities may arise from (relatively local) symmetries present in the $3$D structure (e.g., four similar sides of a clock tower, spherically symmetric dome of a cathedral, etc.) or repeated structures in the scene (e.g., multiple buildings with similar windows). Such \textit{ambiguous} structures may cause unrelated features in different images to be incorrectly matched, and hence result in major errors in structure recovery, e.g.~partial structure estimates incorrectly superimposed on top of each other, \textit{phantom} structures floating in the scene, broken structural integrity due to clustering, etc. As a result, for scenes with ambiguous structures, special care must be taken to disentangle incorrect correspondences of local features. In this section, we shortly discuss some of the recent progress in the literature aiming to handle such ambiguities.

A disambiguation method based on a global integration of geometric relations between images in a Bayesian framework was introduced by~\cite{ZachDisambig}. In order to detect and remove inconsistently matched features, the main idea employed in~\cite{ZachDisambig} is to utilize consistency relations induced by (invertible) pairwise geometric transformations (e.g., homographies, relative camera rotations, similarity transformations between local $3$D reconstructions) along cycles of related images (i.e., images presumably sharing measurements of similar parts of the scene). More concretely, \cite{ZachDisambig} considers the underlying graph of related images (having edges endowed with the pairwise transformations) and proposes to detect inconsistent edges using the fact that (in the noiseless case) chaining all pairwise transformations along any cycle must result in the identity transformation. In order to identify inconsistent edges from cycle consistencies, \cite{ZachDisambig} uses a Bayesian inference approach, where the inconsistency of each edge is represented using a latent binary variable. The objective is then to maximize the joint posterior probability of these latent variables given inconsistencies of each cycle\footnote{Since, in principle, the underlying graph can have an exponentially large number of cycles, \cite{ZachDisambig} restricts the number of inspected cycles to the set of all triangles together with cycles of maximum length six induced by a set of cycle bases (specifically, spanning tree bases), cf.~\S3 in~\cite{ZachDisambig} for further details.} they belong to (i.e., given some measure of the deviations, from the identity transformation, of the transformations corresponding to these variables chained along cycles). In order to obtain an approximate (more precisely, a locally optimal) solution to this difficult problem, \cite{ZachDisambig} uses \textit{loopy belief propagation}. Also, to obtain quality certificates, \cite{ZachDisambig} proposes to approximate the optimization of the log-likelihood function corresponding to the posterior probability using a (standard) convex relaxation approach (cf.~\S3 in~\cite{ZachDisambig} for the details). Lastly, \cite{ZachDisambig} provides experimental evidence of the improvements gained by using their overall disambiguation approach.

Relying solely on pairwise geometric relations for disambiguation may be insufficient for scenes involving large duplicate structures, which result in large, self-consistent sets of incorrectly matched feature pairs. A relatively recent method studied in~\cite{RobertsDuplicates}, which fuses an \textit{expectation maximization} (EM) algorithm to estimate camera poses and a sampling technique (similar to RANSAC~\cite{ransac}) to identify erroneous matches, aims to disambiguate such instances. In designing their method, the basic argument in~\cite{RobertsDuplicates} is that, merely obtaining a geometrically consistent set of edges (as in~\cite{ZachDisambig}) may be insufficient for disambiguation since such an approach inherently assumes \textit{the statistical independence of the erroneous matches}. For camera motion estimation, \cite{RobertsDuplicates} uses a Gaussian mixture model (involving hidden binary variables to represent correct/erroneous matches as in~\cite{ZachDisambig}) to infer the erroneous matches that are \textit{geometrically inconsistent} with the majority of the others. The inference employs an iterative EM algorithm applied first to relative camera rotations, to obtain camera orientations in $\mbox{SO}(3)$, and then to (a spanning tree of) triplet $3$D reconstructions, to obtain a camera motion estimate (cf.~\S3 in~\cite{RobertsDuplicates} for the details). Since such an approach applied to the whole pairwise measurements with an underlying assumption of independence is argued to be ineffective for disambiguation, \cite{RobertsDuplicates} uses a specifically designed \textit{random sampling} of spanning trees of the pairwise measurement graphs, which constitutes the main contribution of~\cite{RobertsDuplicates}. To increase the probability of sampling desired spanning trees (i.e., minimal sets of pairwise measurements with possibly higher accuracy), \cite{RobertsDuplicates} uses an efficient sampling method from the literature for generating random spanning trees according to a specific distribution of spanning trees. The distribution is based on a set of edge weights, where the probability of each tree is proportional to the product of its edge weights. To determine the edge weights, \cite{RobertsDuplicates} considers two sources of information from the images: (1) the ratio of the number of matched features between an image pair to the number of all features (in one of the images in the pair) matched to any other image (adjusted by the distance of the missing correspondences to the matched features between the image pair), and (2) in the presence of \textit{timestamps} for the cameras, the ratio between the time difference of the matched image pair and the smallest time difference of any matched image pair including one of the images in the pair. For each sampled minimal hypothesis, \cite{RobertsDuplicates} computes a set of matches consistent with the hypothesis and updates the estimated camera motion (using the EM approach discussed above, while fixing the edges of the sampled spanning tree to be correct edges), and selects the hypothesis with the highest score to obtain the final camera motion estimate (cf.~\S5 of~\cite{RobertsDuplicates} for the numerical details). A final bundle adjustment is applied, only to the matched pairs inferred as correct, to obtain the refined camera motion and the $3$D reconstruction. Experimental results provided in~\cite{RobertsDuplicates}, using relatively small sets of highly similar images, demonstrate improvements obtained via the disambiguation approach, and also suggest an increase in computational time required to obtain the final results (e.g., disambiguation steps for the CUP dataset of $64$ images and the BLDG dataset of $76$ images take $44$ and $90$ minutes, respectively). 

Another disambiguation method based on the optimization of a measure of missing image projections of potential $3$D structures is introduced in~\cite{SeeingDouble}. To quantify the quality of a $3$D reconstruction, \cite{SeeingDouble} employs the fact that a $3$D point has similar appearances in images it is captured. A SIFT descriptor for a $3$D point is defined in~\cite{SeeingDouble} to be the SIFT descriptor of the corresponding feature point in its \textit{most front parallel image} (determined by using an approximate surface normal computed for the $3$D point). Using this definition, one can hope to identify the visibility of a $3$D point in an image by matching its SIFT descriptor (as defined above) to the SIFT descriptor of its projection on the image. Hence, the quality of a $3$D reconstruction can be quantified by measuring the average probability $P_{miss}(p,i)$ (averaged over images and $3$D points) of its points $p$ to be missing in the images $i$, i.e.~the average probability that the SIFT descriptors of its $3$D points do not match with those of their image projections (cf.~\S3 in~\cite{SeeingDouble} for parametric assumptions~\cite{SeeingDouble} makes to compute $P_{miss}(p,i)$). To minimize the average $P_{miss}(p,i)$, \cite{SeeingDouble} first computes pairwise $3$D reconstructions and relative camera poses. The optimization is performed on spanning trees of the match graph (induced by edges corresponding to camera pairs with a computed relative pose), where~\cite{SeeingDouble} employs various heuristics in each iteration to improve the efficiency of its search (cf.~\S5 in~\cite{SeeingDouble} for the details). Lastly,~\cite{SeeingDouble} concludes with experimental results demonstrating the accuracy, efficiency and limitations of its disambiguation approach. 

Most of the disambiguation methods in the literature are designed to operate on relatively small sets of images designed to exemplify relatively difficult instances of the SfM disambiguation problem. However, the level of difficulty and the efficiency requirements of large, unordered image collections can be significantly different. A recent method, purely based on a graph theoretic approach, designed to disambiguate such large datasets is studied in~\cite{WilsonSnavelyDisambig}. The main concept in~\cite{WilsonSnavelyDisambig} is the bipartite \textit{visibility graph} $G = (I,T,E)$, where nodes are the images $I$ and the tracks (i.e., sets of matched features across several images) $T$, and an edge $(i, t)\in E$ exists if the track $t$ includes a feature in image $i$. The main idea in~\cite{WilsonSnavelyDisambig} is to identify and remove \textit{bad tracks} in $T$ (i.e.~tracks which correspond to more than one $3$D points, possibly due to ambiguities in the scene), which tend to form \textit{bridges} between unrelated parts of the scene (and, hence, between the corresponding images and tracks of these scenes in $G$). To quantify the quality of a track $t$ (i.e., its dissimilarity to a bridge-type node), \cite{WilsonSnavelyDisambig} uses the \textit{bipartite local clustering coefficient} of $t$ defined to be the ratio of the number of closed $4$-paths centered at $t$ to the number of all $4$-paths centered at $t$. Also, in order to eliminate the sensitivity of bipartite local clustering coefficients to the variations in the sizes of local clusters in the visibility graph (e.g., resulting from largely varying popularities of different parts of a touristic scene), \cite{WilsonSnavelyDisambig} computes a covering subgraph $G'$ of the visibility graph $G$ with improved uniformity (cf.~\S5 in~\cite{WilsonSnavelyDisambig} for technical details and specific parameters). The bad tracks of $G'$ are determined to be those having bipartite local clustering coefficients less than a threshold, and are removed from $G'$ (the same procedure is applied also to $G$). The remaining (good) tracks and their corresponding images are used in an incremental SfM pipeline from~\cite{SnavelyRome} to obtain an initial (skeletal) reconstruction, and then remaining tracks and images in $G$ are used to enrich the reconstruction. Although this purely graph theoretic approach may be insufficient for difficult SfM disambiguation instances, \cite{WilsonSnavelyDisambig} provides extensive empirical results (for large Internet photo collections) demonstrating the accuracy and the high computational efficiency of their disambiguation technique.

For alternative approaches aiming to resolve the difficulties in SfM for symmetric and ambiguous scenes, we refer the reader to other sources, e.g.~\cite{CohenSymmetries,SnavelyModel,SchaffalitzkyDisambig,MartinecRotations,PDM}.

\subsection{Alternative Features and Camera Models}
Our discussion in the previous sections has mainly focused on SfM techniques based on \textit{point features} extracted from images formed by \textit{perspective projection}. In this section, we touch upon some of the SfM methods in the literature designed for alternative features and camera models.

A relatively popular alternative to using correspondences of point features for SfM is to use corresponding \textit{lines} across multiple images. A solution to a line-based SfM instance is composed of a camera motion estimate and $3$D line reconstructions, which may be more suitable, as compared to $3$D point reconstructions, for applications like modeling of urban scenes, augmented reality, etc. A crucial point for line-based SfM methods is the mathematical representation of the $4$-dimensional set of $3$D lines and a corresponding model for their measurements in images, since the usefulness of an existing representation may depend on the particular problem considered. Relatively early factorization methods for line-based SfM (cf., e.g., \cite{LineBasedFactor1,LineBasedFactor2}) are usually constrained in their accuracy and applicability to scenes with occlusions. A direct reprojection error minimization approach (susceptible to local minima) is proposed in~\cite{LineBased3}, where a line $l$ is represented as $l = (\mathbf{d},\mathbf{v})$ using a \textit{closest point $\mathbf{d}$ to the origin} and a unit norm \textit{direction} $\mathbf{v}$\footnote{Note that this representation is a two-to-one correspondence, between the $4$-dimensional manifold of parameters $(\mathbf{d},\mathbf{v})$ satisfying $\|\mathbf{v}\|=1, \mathbf{v}^T\mathbf{d} = 0$ and the set of $3$D lines, since $(\mathbf{d},\mathbf{v})$ and $(\mathbf{d},-\mathbf{v})$ correspond to the same line.}. Two different representations of lines are used in~\cite{LineBased2} for the two problems of triangulation of matched features and bundle adjustment. For triangulation, Pl\"{u}cker coordinates are used to represent lines, mainly due to the resulting linearity of the reprojected lines as functions of the $3$D lines. For a pair of $3$D points $\mathbf{x} = [\bar{\mathbf{x}} \ x_0]$ and $\mathbf{y} = [\bar{\mathbf{y}} \ y_0]$ in homogeneous coordinates, the $6$-dimensional vector $[\mathbf{a} \ \mathbf{b}]$ representing the Pl\"{u}cker coordinates of the line through the points $\mathbf{x}$ and $\mathbf{y}$ satisfies $\mathbf{a} = \bar{\mathbf{x}}\times\bar{\mathbf{y}}$, $\mathbf{b} = x_0\bar{\mathbf{y}}-y_0\bar{\mathbf{x}}$ and $\mathbf{a}^T\mathbf{b} = 0$. A linear and two quasi-linear algorithms (combined with a final step of rounding to the Pl\"{u}cker coordinates) for triangulation of line features are introduced in~\cite{LineBased2}. For bundle adjustment, \cite{LineBased2} employs \textit{the orthonormal representation} of $3$D lines $l = (U,W)$, for $U\in\mbox{SO}(3)$ and $W\in\mbox{SO}(2)$ (cf.~\S5.3 in~\cite{LineBased2} for the details of the computation of $U,W$). This is mainly due to the completeness and minimal updatability (i.e., only $4$ parameters are required to update the representation) of the orthonormal representation, which also lacks gauge freedoms and internal constraints, making it a numerically stable choice for the nonlinear, unconstrained bundle adjustment step. Experimental results are also provided in~\cite{LineBased2} to evaluate the performance of their overall approach. An alternative method specifically designed for line-based reconstruction of $3$D urban scenes is introduced in~\cite{LineBased1}. The main idea in~\cite{LineBased1} is to classify the lines in the images into different orientation classes (e.g., horizontal lines, vertical lines, etc.) using an expectation-maximization approach to estimate $3$D vanishing point assignments (representing vanishing directions, at least $3$ of which are assumed to be mutually orthogonal) for every pixel in the images. Such a classification is argued in~\cite{LineBased1} to improve the accuracy of matching of lines between images (since only lines in separate classes are not allowed to be matched), and also increase the efficiency and accuracy of the $3$D structure reconstruction by reducing the degrees of freedom of $3$D lines during optimization. To represent a $3$D line, \cite{LineBased1} considers an embedding of the $4$-dimensional manifold of lines in the eleven dimensional space $\mbox{SO}(3)\times\R^2$, where the $\mbox{SO}(3)$ and $\R^2$ parts essentially represent plane of orthogonality of the line and the point of intersection of the line and the plane of orthogonality, respectively. In order to overcome the singularities induced by this over parametrization during non-linear optimization for $3$D structure recovery, \cite{LineBased1} uses local parameterizations around each point in $\mbox{SO}(3)\times\R^2$ (encoded by the exponential map of the Lie group $\mbox{SO}(3)$) with varying degrees of freedom for each class of lines (e.g., horizontal lines have $3$ degrees of freedom, whereas vertical lines have only $2$ degrees of freedom). Experimental results providing the $3$D line reconstructions (for relatively small image sets) are also provided in~\cite{LineBased1}. For additional reading on urban scene reconstruction, we refer the reader, e.g., to~\cite{Urban3DSurvey,LinesFromVideo} and the references therein. 

\begin{figure}[!htbp]
\begin{center}
\includegraphics[trim=5cm 3.5cm 6cm 5cm, clip=true, width=0.7\linewidth]{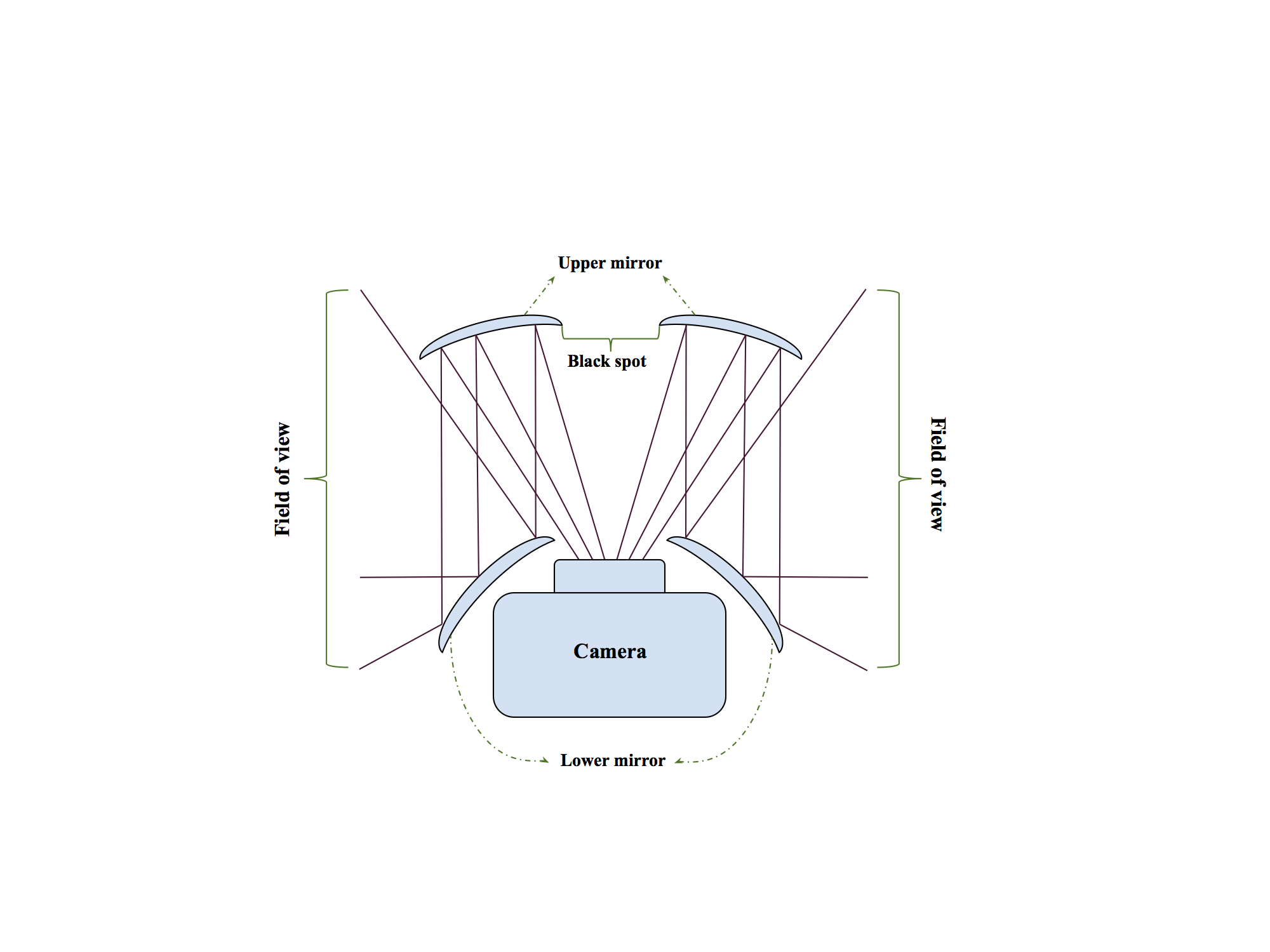}
\end{center}
\caption{A (simplified) diagram for a 2D slice of an omnidirectional camera with two mirrors (upper and lower).\label{fig:OmniDiagram}}
\end{figure}

There exist several works in the SfM literature based on relatively uncommon camera models that aim to solve the SfM problem with specific applications in mind or improve the accuracy of existing $3$D recovery techniques by using different measurement apparatus (i.e.~different cameras). One such study, considering the SfM problem for \textit{omnidirectional cameras} (i.e., cameras with large visual coverage, possibly having a $360^{\circ}$ field of view in the horizontal plane, or one that approximately covers the entire visual sphere, cf.~Figure~\ref{fig:OmniDiagram} for a simplified diagram), is given in~\cite{OmniSfM}. 
The epipolar constraints for (catadioptric) omnidirectional cameras are derived in~\cite{OmniSfM}, and a two step minimization of the error in these constraints, composed of an initial (classical) estimation of the relative motion parameters from the essential matrix estimates and a direct minimization of the epipolar errors with respect to the relative motion parameters that takes the solution of the first step as an initial point, is employed to estimate the pairwise camera motions. A (brief) analysis to quantify the errors in the estimated motion parameters induced by the uncertainties in the correspondence measurements, and also experimental results demonstrating (occasional) improvements in accuracy obtained by using omnidirectional cameras instead of the conventional ones, are also provided in~\cite{OmniSfM}. A more systematic and detailed study of SfM from variants of omnidirectional cameras (i.e., fish-eye lenses and catadioptric cameras with conical mirrors) is provided in~\cite{MicusikWideCam}. The two-view geometry estimation, camera calibration and $3$D reconstruction problems are all studied in~\cite{MicusikWideCam}. Different parametric measurement models are introduced in~\cite{MicusikWideCam} for para-catadioptric cameras and fish-eye lenses, and the corresponding camera auto-calibration problems are solved by minimizing an angular error measure via \textit{polynomial eigenvalue} computation. For other works on SfM in the omnidirectional setup, cf., e.g., \cite{OmniExtra1,OmniExtra2,GenericExtra1,OmniExtra3} and the references therein. An alternative approach in the literature is to use \textit{generic camera models}, representing cameras as sets of projection rays, to incorporate various kinds of cameras (e.g., cameras with radial distortion, non-central catadioptric cameras, pinhole cameras, stereo rig cameras, etc.) in a unified framework. Interestingly, such a generic SfM framework allows simultaneous processing of images captured by different camera types. The main approach for motion estimation in the generic setting is to assign projection rays to interest points in images, by using camera calibration information, and to minimize the error in the \textit{generalized epipolar constraints} (also known as \textit{Pless equations}) written in terms of the corresponding projection rays (e.g., represented using Pl\"{u}cker coordinates). Also, $3$D structure points can be estimated using the geometric relations between the projection rays and the $3$D interest points (cf., e.g., \S3 and \S4 in~\cite{GenericExtra2}). For experimental results and technical details of these generic SfM methods, we refer the reader to~\cite{GenericExtra2,MouragnonGeneric,GenericExtra1} and the references therein.

\subsection{Popular Software Packages and Data Sources}
For most of the existing SfM methods, the most computationally involved part is the bundle adjustment step. As a result, efficient and accurate implementations for bundle adjustment are crucial for high quality SfM solutions. One of the most popular software packages for bundle adjustment is the \textit{bundler} package of~\cite{SnavelyData}. Essentially, bundler is based on an incremental application of a modified version of the \textit{sparse bundle adjustment} (SBA) package of~\cite{SBA}, and, being an incremental method, it incorporates a few images at a time to produce a camera motion and a $3$D structure estimate from a set of images, image features, and image matches provided as input. In general, for large, unordered, weakly connected sets of images, incremental methods may suffer from accumulation of estimation errors, which result in large deformations in the structure estimate. Additionally, several intermediate applications of bundle adjustment may hinder the overall computational efficiency. Nevertheless, the bundler package has been empirically shown to produce accurate solutions that are usually considered as reference solutions for experimental comparisons (cf., e.g., \cite{LUD,Snavely1D}). The efficiency of the crucial SBA method of~\cite{SBA} at the core of bundler essentially arises from (as the name obviously suggests) the exploitation of the sparsity of the Jacobian in the normal equations (\ref{eq:BANormalEqs}) of the Levenberg-Marquardt iterations. We note that, as a standalone package independent of bundler, SBA is an efficient and stable alternative for final refinements of initial motion and structure estimates, i.e. for a final bundle adjustment step. 

A highly efficient multicore CPU and GPU implementation of the preconditioned conjugate gradients (CG) solver (similar to~\cite{SnavelyBALarge}, cf.~\S\ref{sec:StructEst}) for bundle adjustment, named \textit{multicore bundle adjustment}, was introduced in~\cite{PBA}. The multicore bundle adjustment methodology is reported in~\cite{PBA} to maintain high accuracy (i.e., comparable to alternatives) while requiring less memory usage and providing runtime improvements of up to ten times for the CPU, and up to thirty times for the GPU implementations. This improvement in efficiency is solely based on the optimized usage of parallel computational resources (i.e., multicore CPUs and GPUs). An incremental SfM package, named \textit{VisualSfM}, based on the multicore bundle adjustment algorithm of~\cite{PBA}, was later provided in~\cite{VisualSfM}. Empowered with a very useful graphical user interface (GUI) that allows the explicit visualization of several different stages of SfM, VisualSfM can be considered to be an efficient and relatively stable incremental SfM solver (though, possessing the difficulties inherent in the incremental approach). Additionally, by the integration of the GPU implementation of SIFT given in~\cite{SiftGPU}, VisualSfM can be used as an end-to-end SfM engine taking as input a set of images and producing as output a motion and a $3$D structure estimate (however, such a \textit{black box} usage may result in degradation of accuracy in the final estimates).

\begin{figure}[!htbp]
\centering
\subfloat[]{\includegraphics[trim=11.5cm 7cm 13cm 1cm, clip=true, width=0.45\linewidth]{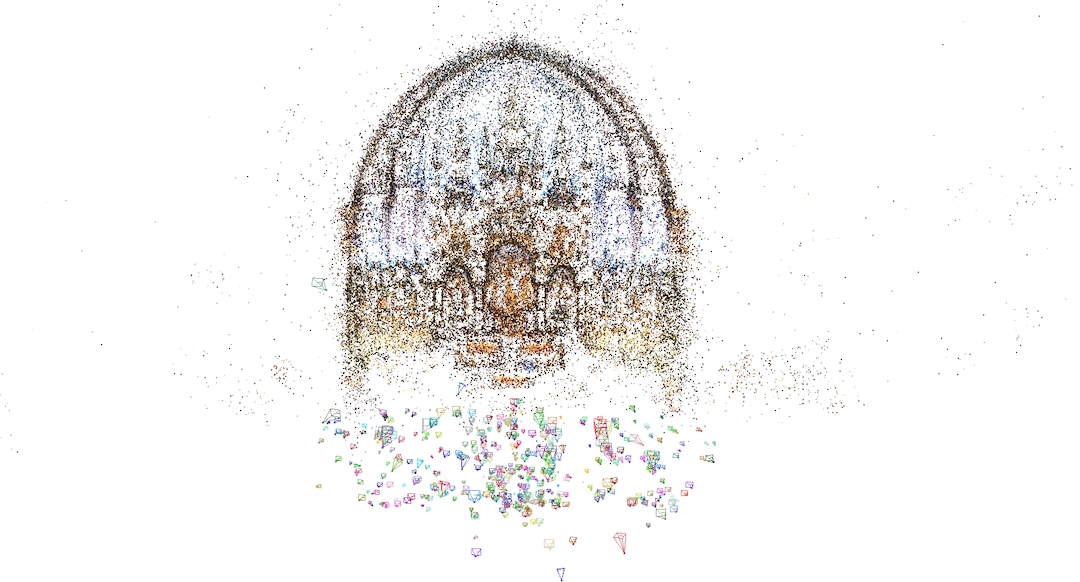}} \hspace{0.1in}
\subfloat[]{\includegraphics[trim=5.5cm 0.5cm 6cm 0cm, clip=true, width=0.50\linewidth]{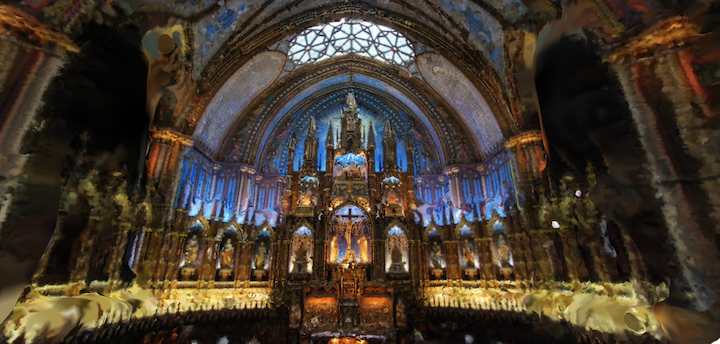}}
\caption{3D structures for the Montreal Notre Dame dataset from~{\rm\cite{Snavely1D}}: {\rm(a)} Sparse 3D structure obtained by using the method of~{\rm\cite{LUD}} for camera motion estimation and the PBA package of~{\rm\cite{PBA}} for final refinement. {\rm(b)} A snapshot of the dense 3D reconstruction obtained by using the PMVS package of~{\rm\cite{Dense3D,PMVS}} with the motion estimate of~{\rm\cite{LUD}}. \label{fig:SparseVSDense3D}}
\end{figure}

Estimation of a $3$D structure in SfM usually refers to the computation of a \textit{point cloud} (also known as the \textit{sparse structure}) corresponding to points from the stationary scene. However, it can also be desired to produce a \textit{dense structure}, e.g.~for purposes of visualization, $3$D printing, etc., corresponding to a relatively continuous representation of the corresponding scene, in the form of a dense set of rectangular patches covering the surfaces visible in the input images (cf.~Figure~\ref{fig:SparseVSDense3D} for an example). 
This problem is known as \textit{multi-view stereopsis}. An accurate algorithm for this purpose, namely \textit{patch-based multi-view stereo software} (PMVS), which is based on a three-step procedure of matching, expansion, and filtering of patches of keypoints\footnote{Here, we do not provide any detailed description of this particular algorithm for the optional procedure of dense $3$D reconstruction, please see~\cite{Dense3D,PMVS} for details.}, is given in~\cite{Dense3D,PMVS}. When provided with accurate initial camera motion and calibration estimates, PMVS produces high quality, well-connected dense $3$D reconstructions. However, dense reconstruction can be computationally very expensive for large sets of images. To tame this computational inefficiency, \cite{FurukawaPartial,CMVS} proposed a highly scalable (to tens of thousands of images) distributed approach based on the decomposition of the images into overlapping sets to be processed in parallel (i.e., to produce local dense $3$D structures using, e.g., PMVS). The emerging dense structures are then merged to produce the final dense $3$D estimate. The empirical results of~\cite{FurukawaPartial} demonstrate the high accuracy and improved efficiency of this distributed approach. We also note that the software packages PMVS and CMVS are integrated into the VisualSfM package.

A recent end-to-end library, named \textit{Theia Vision Library}, which incorporates efficient, reliable and supported implementations of various alternative methods for each stage of SfM, is available in~\cite{Theia}. Theia provides implementations of various methods and useful routines for image handling, feature detection, description and matching, relative pairwise pose estimation, camera modeling, global camera orientation and location estimation, initial structure estimation, incremental SfM, motion and structure refinement using bundle adjustment, and gauge symmetry computation for performance evaluation. Theia also provides extensive documentation for all of its components and is open to contributions for extensions and improvements. 

Evaluation of the accuracy and the efficiency of many of the existing methods for SfM are solely based on empirical observations. Consequently, easily accessible datasets (i.e., sets of raw images, extracted and matched features, intrinsic camera calibrations, etc.) possessing specific characteristics, ranging from benchmark image sets with ground truth motion parameters to large, unordered Internet photo collections, are of great value for comparative performance evaluations. Additionally, and perhaps more importantly, collection and preprocessing (e.g., feature extraction and matching) of such datasets can be quite time consuming and expensive. Hence, the unavailability of such datasets, may significantly hinder the development and testing of new methods for SfM. A relatively early work providing several benchmark datasets (now considered to have small sizes) including high resolution images, intrinsic camera parameters, ground truth camera motion and resulting dense $3$D constructions, is~\cite{FountainData} (datasets are available at \url{http://cvlabwww.epfl.ch/data/multiview/}). The availability of ground truth intrinsic and extrinsic camera parameters in these datasets makes them useful for evaluating camera motion recovery accuracy (cf., e.g.,~\cite{MicaAmitSfM,CvXSfM}) with \textit{high precision}. As discussed in\S\ref{sec:StructEst}, more recently, there has been a growing interest in SfM instances involving large and unordered images. A highly popular source of data, including large SfM instances with original images, extracted and matched features, relative camera poses, intrinsic camera parameters, etc., based on the works~\cite{SnavelyBALarge,SnavelyRome,SnavelyData,WilsonSnavelyDisambig,Snavely1D} and more, is the \textit{BigSFM} platform accessible at \url{http://www.cs.cornell.edu/projects/bigsfm/} (also see the links at the data section of the BigSFM webpage). As a widely used source of data, the BigSFM platform provides a (relatively) standardized basis for comparative performance evaluation of various methods in the literature targeting efficient feature extraction and matching, global camera motion estimation, incremental and global bundle adjustment, handling symmetries and ambiguities, geo-positioning, etc.

\section{Conclusion}
\label{sec:Conclusion}
Our survey covered various methods in the SfM literature pertaining to camera location estimation, $3$D structure and motion refinement, simultaneous localization and mapping (SLAM), feature extraction and matching, symmetric and ambiguous scenes, and alternative features and camera models. As also mentioned in the previous sections, for additional topics and methods in the literature, we refer the reader to other works, e.g.~\cite{HartleyBook,BundleAdjustment,OliensisCritique,EpipolarReview,TronSurvey,KanadeFactorization,Urban3DSurvey,GLOH,DescSurvey2}. In this section, we conclude by briefly pointing to potential areas of progress in the SfM problem.

For most of the existing methods in the SfM literature, one of the most computationally intensive steps (usually assumed to have been accomplished by an existing method) is the feature extraction and matching. Even though there exist sufficiently accurate methods for this task, the computational cost of these methods can be prohibitive for specific applications, and some of the relatively efficient techniques turn out to induce significant degradation in accuracy. In other words, the existing methods seem to have saturated in improving the computational efficiency without suffering losses in accuracy. In this respect, more efficient feature extraction and matching methods, which have improved invariance properties and enable fusion of local and global image characteristics (e.g., to maintain higher resilience to ambiguities in images) are of high value.

As discussed in \S\ref{subsec:Ambiguities}, existing SfM methods for ambiguous scenes can be classified into the two main categories of relatively computationally inefficient methods designed for high proportions of mismatched features (usually observed for small, experimental image sets) and relatively efficient techniques for large image sets (e.g., community photo collections) that tend to fail in disambiguation of scenes with large ambiguous structures. Hence, we believe that there is potential for alternative methods with acceptable levels of computational efficiency that have improved robustness to higher numbers of mismatched features. 

Considering the recent boost of interest in $3$D scene perception and reconstruction (e.g., for augmented and virtual reality applications, self-driving cars, etc.), we believe that specific SfM algorithms targeting efficient solutions of relatively tightly constrained problem instances (e.g., accurate and fast depth estimation with known camera motion) will attract more attention in the future. Consequently, design and analysis of efficient SfM techniques for such instances, which admit predictable stability in accuracy and provable robustness, is an important potential direction of progress. 

Finally, even though there exist various theoretical works in the literature that study fundamental problems in SfM and/or provide rigorous analysis of stability and robustness of specific methods (cf., e.g., the early works~\cite{ChiusoSoattoOptimal,MaThy,SoattoVisualMotion} and the relatively recent works~\cite{CvXSfM,ShapeFit,ShapeFitJoint}), we believe that the SfM community would still highly benefit from rigorous results on fundamental problems (e.g., what is the theoretically maximal amount of mismatched features or level of noise in the images that can be tolerated for a stable structure recovery, and can this be achieved efficiently?) and theoretical analysis of stability, robustness and computational efficiency of existing or new methods.\\

\noindent {\bf Acknowledgements.} \ \ The authors wish to thank Noah Snavely, David G. Lowe and Federica Arrigoni for their permission to include  Figures~\ref{fig:Snavely1DSfM},\ref{fig:SnavelyBALarge},\ref{fig:SnavelySkeletal},\ref{fig:SnavelyPhotoTourism}, Figures~\ref{fig:SIFTDoG},\ref{fig:SIFTDescriptor} and Figure~\ref{fig:GSPdiag}, respectively (see the captions of these figures for the corresponding publications of these authors). A.S.~was partially supported by Award Number R01GM090200 from the NIGMS, FA9550-12-1-0317 from AFOSR, Simons Foundation Investigator Award and Simons Collaboration on Algorithms and Geometry, and the Moore Foundation Data-Driven Discovery Investigator Award. R.B.~was supported in part by the Israel Science Foundation grant No.~1265/14 and by the Minerva Foundation with funding from the Federal German Ministry for Education and Research.

\newpage
\bibliographystyle{abbrv}
\bibliography{SfMbib}

\end{document}